\documentclass[10pt,twocolumn,letterpaper]{article}

\usepackage{cvpr}              

\usepackage{graphicx}
\usepackage{amsmath}
\usepackage{amssymb}
\usepackage{booktabs}

\usepackage{times}
\usepackage{epsfig}
\usepackage{algorithm}
\usepackage{algorithmic}

\usepackage{diagbox}
\usepackage{stackengine}
\usepackage{color}
\usepackage[dvipsnames]{xcolor}

\newcommand\ChangeRT[1]{\noalign{\hrule height #1}}
\newcommand\xrowht[2][0]{\addstackgap[.5\dimexpr#2\relax]{\vphantom{#1}}}

\usepackage{listings}
\usepackage{xcolor}
\definecolor{codegreen}{rgb}{0,0.6,0}
\definecolor{codegray}{rgb}{0.5,0.5,0.5}
\definecolor{codepurple}{rgb}{0.58,0,0.82}
\definecolor{backcolour}{rgb}{0.95,0.95,0.92}
\lstdefinestyle{mystyle}{
    backgroundcolor=\color{backcolour},   
    commentstyle=\color{codegreen},
    keywordstyle=\color{magenta},
    numberstyle=\tiny\color{codegray},
    stringstyle=\color{codepurple},
    basicstyle=\ttfamily\footnotesize,
    breakatwhitespace=false,         
    breaklines=true,                 
    captionpos=b,                    
    keepspaces=true,                 
    numbers=left,                    
    numbersep=5pt,                  
    showspaces=false,                
    showstringspaces=false,
    showtabs=false,                  
    tabsize=2
}
\usepackage[pagebackref,breaklinks,colorlinks]{hyperref}

\usepackage[capitalize]{cleveref}
\crefname{section}{Sec.}{Secs.}
\Crefname{section}{Section}{Sections}
\Crefname{table}{Table}{Tables}
\crefname{table}{Tab.}{Tabs.}

\def\cvprPaperID{599} 
\def\confName{CVPR}
\def\confYear{2022}

\begin{document}

\title{SimT: Handling Open-set Noise for Domain Adaptive Semantic Segmentation}

\author{Xiaoqing Guo$^{1}$~~~~~~~~~~~~Jie Liu$^{1}$~~~~~~~~~~~~Tongliang Liu$^{2}$~~~~~~~~~~~~Yixuan Yuan$^{1}$\thanks{Yixuan Yuan is the corresponding author. \newline $~~~~~~$This work was supported by Hong Kong Research Grants Council (RGC) General Research Fund 11211221(CityU 9043152).}
\\
$^{1}$City University of Hong Kong~~~~~~~~~~~~~~$^{2}$University of Sydney\\
{\tt\small \{xqguo.ee, jliu.ee\}@my.cityu.edu.hk}~~~~~{\tt\small tongliang.liu@sydney.edu.au}~~~~~{\tt\small yxyuan.ee@cityu.edu.hk} 
}

\maketitle

\begin{abstract}
This paper studies a practical domain adaptive (DA) semantic segmentation problem where only pseudo-labeled target data is accessible through a black-box model. Due to the domain gap and label shift between two domains, pseudo-labeled target data contains mixed closed-set and open-set label noises. In this paper, we propose a simplex noise transition matrix (SimT) to model the mixed noise distributions in DA semantic segmentation and formulate the problem as estimation of SimT. By exploiting computational geometry analysis and properties of segmentation, we design three complementary regularizers, \ie volume regularization, anchor guidance, convex guarantee, to approximate the true SimT. Specifically, volume regularization minimizes the volume of simplex formed by rows of the non-square SimT, which ensures outputs of segmentation model to fit into the ground truth label distribution. To compensate for the lack of open-set knowledge, anchor guidance and convex guarantee are devised to facilitate the modeling of open-set noise distribution and enhance the discriminative feature learning among closed-set and open-set classes. The estimated SimT is further utilized to correct noise issues in pseudo labels and promote the generalization ability of segmentation model on target domain data. Extensive experimental results demonstrate that the proposed SimT can be flexibly plugged into existing DA methods to boost the performance. The source code is available at \url{https://github.com/CityU-AIM-Group/SimT}.
\end{abstract}
\section{Introduction}
\label{sec:intro}
DA semantic segmentation aims to adapt a segmentation model for the target domain through transferring knowledge from a source domain \cite{liu2021bapa}. Considering the inefficient transmission and privacy issues in source domain knowledge, effective adaptation with only pseudo-labeled target data obtained through a black-box model is practical in real-world deployments \cite{liang2021distill}. This paper mainly focuses on this realistic scenario, and the challenge of this DA semantic segmentation problem lies in noisy pseudo labels generated for target domain data. To mitigate noisy adaptation, some methods leverage confidence score \cite{iqbal2020mlsl, pan2020unsupervised, zou2018unsupervised} or uncertainty \cite{liang2019exploring, zheng2021rectifying} to select reliable pseudo labels. For example, prior work \cite{zou2018unsupervised} defines a threshold to eliminate low-confidence pseudo-labeled pixels. However, simply dropping \textit{confusing pixels} with highly potential noise labels might lead to learn from a biased data distribution \cite{zhu2021second}. In fact, confusing pixels are more prone to be mislabeled and critical for accurate predictions, hence need to be carefully addressed. Instead of discarding unreliable pseudo-labeled target data, other methods denoise pseudo labels based on prototypes \cite{zhang2021prototypical} or correct supervision signal through noise transition matrix (NTM) \cite{guo2021metacorrection}. Nevertheless, these approaches heavily rely on the assumption of a shared label set between two domains and only consider the closed-set noisy pseudo labels. In real-world scenarios, the label space of target domain may not be consistent with that of source domain, and \textit{open-set} class instances are commonly encountered in target domain. Hence, enhancing the segmentation performance on target domain needs to carefully address \textit{confusing pixels} and \textit{open-set label noises}. 

\begin{figure}[t]
\centering
\includegraphics[width=82mm]{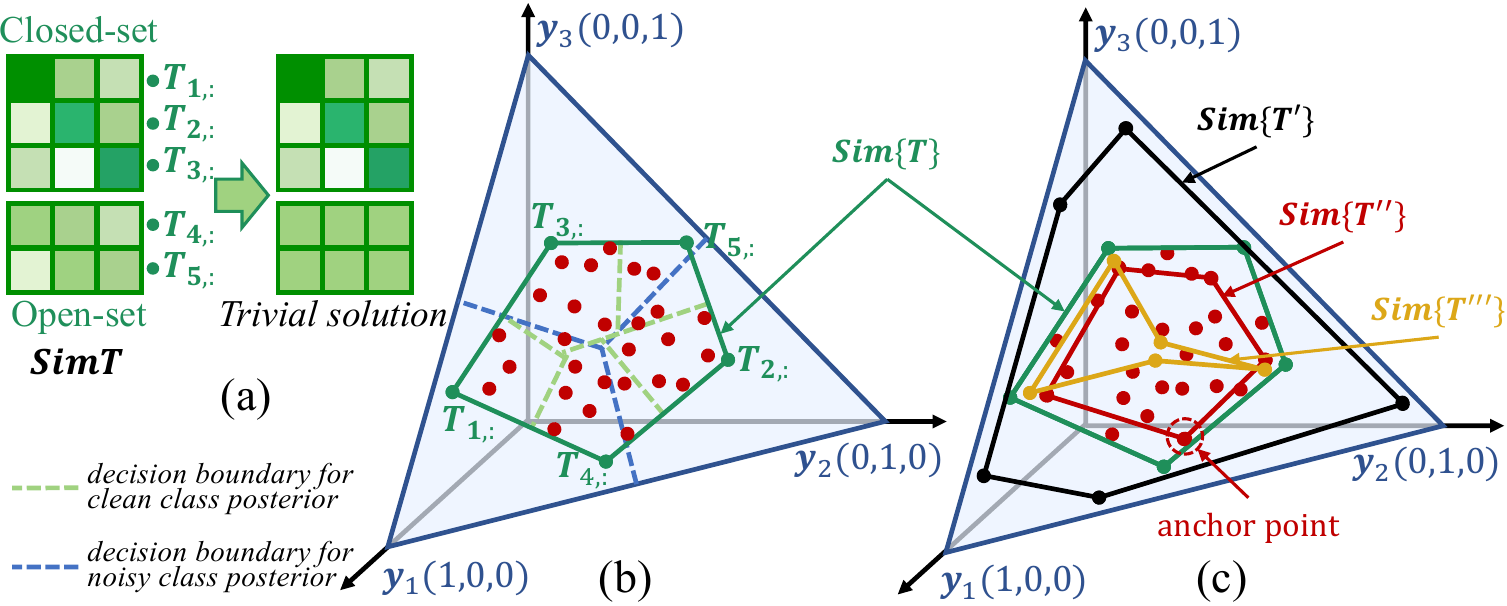}
\caption{Geometric illustration of SimT with the setting of 3 pseudo label classes and 5 ground truth label classes. Note that coordinates $(x, y, z)$ of any points in \textcolor{RoyalBlue}{blue triangle} satisfy conditions of $x+y+z=1$  and $0 < x,y,z<1$.}
\label{fig:Motivation}
\end{figure}

With the aforementioned aims, we model the closed-set and open-set noise distributions in target pseudo labels through a SimT, which is a class-dependent and {instance-independent} transition matrix\footnote{The complex instance-dependent label noise in real-world can be well approximated by class-dependent label noise when the noise rate is low \cite{cheng2020learning}.}. Then the modeled noise distribution is leveraged to rectify supervision signals (\ie loss) derived from noisy labels, thus all target samples are fully utilized. Therefore, \textit{the problem of alleviating closed-set and open-set noise issues can be treated as the problem of estimating SimT}. To give an intuitive explanation of SimT, we interpret SimT through a geometric analysis on a toy example in Figure \ref{fig:Motivation}. In this toy example, SimT is defined as a matrix $\boldsymbol{T} \in [0, 1]^{5\times 3}$, as illustrated in Figure \ref{fig:Motivation} (a). Each row, \ie $\boldsymbol{T}_{c,:}$, indicates the probability of label $c$ flipping to three classes, which can be regarded as a $three$-dimensional point within the \textcolor{RoyalBlue}{blue triangle} of Figure \ref{fig:Motivation} (b) since coordinates of this point add up to 1. The first three rows $[\boldsymbol{T}_{1,:}, \boldsymbol{T}_{2,:}, \boldsymbol{T}_{3,:}]$ of SimT model closed-set noise distribution, and the rest rows $[\boldsymbol{T}_{4,:}, \boldsymbol{T}_{5,:}]$ model open-set noise distribution. Given an input instance $\boldsymbol{x}$, the noisy class posterior probability $p(\widetilde{\boldsymbol{y}} \mid \boldsymbol{x})=[p(\widetilde{y}=1 \mid \boldsymbol{x}), ~p(\widetilde{y}=2 \mid \boldsymbol{x}), ~p(\widetilde{y}=3 \mid \boldsymbol{x})]$ can be regarded a \textcolor{Red}{red point} scattered in the \textcolor{RoyalBlue}{blue triangle}. Since SimT bridges noisy class posterior and clean class posterior $p(\boldsymbol{y} \mid \boldsymbol{x})=[p(y=1 \mid \boldsymbol{x}), ~p(y=2 \mid \boldsymbol{x}), ..., ~p(y=5 \mid \boldsymbol{x})]$ through $p(\widetilde{\boldsymbol{y}} \mid \boldsymbol{x})= p({\boldsymbol{y}} \mid \boldsymbol{x}) ~ \boldsymbol{T}$, the noisy class posterior probability $p(\widetilde{\boldsymbol{y}} \mid \boldsymbol{x})$ is interpreted as a convex combination among rows of $\boldsymbol{T}$ with the factor of $p({\boldsymbol{y}} \mid \boldsymbol{x})$. In the $three$-dimensional space, the closer $p(\widetilde{\boldsymbol{y}} \mid \boldsymbol{x})$ is to the vertex $\boldsymbol{T}_{c,:}$ of \textcolor{Green}{green pentagon $Sim\{ \boldsymbol{T} \}$}, the more likely the input $\boldsymbol{x}$ belongs to class $c$. Hence, the simplex formed by rows of $\boldsymbol{T}$ should enclose $p(\widetilde{\boldsymbol{y}} \mid \boldsymbol{x})$ for any $\boldsymbol{x}$. However, since there exists an infinite number of simplices enclosing $p(\widetilde{\boldsymbol{y}} \mid \boldsymbol{x})$, SimT is not identifiable if no assumption has been made. 

A feasible solution to estimate SimT is volume minimization based on the sufficiently scattered assumption where class posterior distribution is far from uniform \cite{li2021provably}. As demonstrated in \cite{li2021provably}, finding the minimum-volume simplex enclosing $p(\widetilde{\boldsymbol{y}} \mid \boldsymbol{x})$ recovers the ground truth NTM and $p(\boldsymbol{y} \mid \boldsymbol{x})$. In semantic segmentation, interior pixels away from object edge usually exhibit high confidence for classification. Those confident instances show discriminative distribution and thus have class posteriors far from uniform. In this regards, the sufficiently scattered assumption is easy to satisfy in segmentation, hence we follow the idea of volume minimization to estimate SimT. Nevertheless, NTM in \cite{li2021provably} only models closed-set noisy labels, and directly applying volume minimization that constrains a square NTM to our SimT is problematic with two main reasons. Firstly, SimT is a non-square matrix, whose volume cannot be calculated through its determination. Secondly, the success of volume minimization depends on the diagonally dominant guarantee of the square matrix, while the diagonally dominant assumption is not available for the open-set part of SimT. Hence, by minimizing volume of SimT, open-set part of SimT will collapse to a trivial solution (Figure \ref{fig:Motivation} (a) and \textcolor{Dandelion}{$Sim\{ \boldsymbol{T'''} \}$} in (c)), leading to undiscriminating open-set feature distributions and non-convex problem. These motivate us to design a computational geometry based algorithm which can detect open-set noises from pseudo-labeled target data and approximate the true SimT.

To this end, we propose three regularizers, \ie \textit{volume regularization}, \textit{anchor guidance} and \textit{convex guarantee}, for the estimation of SimT. Specifically, \textit{volume regularization} for non-square SimT estimation is designed to approximate the true SimT, which is the one with minimum volume (\textcolor{Green}{$Sim\{ \boldsymbol{T} \}$}) among all simplices enclosing $p(\widetilde{\boldsymbol{y}} \mid \boldsymbol{x})$ (\eg, \textcolor{black}{$Sim\{ \boldsymbol{T'} \}$}) in Figure \ref{fig:Motivation} (c). The minimum volume of SimT ensures outputs of segmentation model to fit into clean class posteriors. To prevent the trivial solution of open-set part in SimT, we devise \textit{anchor guidance} regularization. In particular, we first define \textcolor{Red}{anchor point} as the most representative pixel of each class with the largest clean class posterior probability as in Figure \ref{fig:Motivation} (c). \textcolor{Red}{$Sim\{\boldsymbol{T''} \}$} formed by all anchor points is an approximation to the true SimT, hence we leverage anchor points to guide the estimation of SimT. To further facilitate the detection of anchor points in open-set classes, an auxiliary loss enhancing the discriminative feature learning among closed-set and open-set data is designed. Moreover, the trivial solution of open-set part may lead to non-convex \textcolor{Dandelion}{$Sim\{ \boldsymbol{T'''} \}$}. Hence, we advance \textit{convex guarantee} to preserve the convex property of SimT, and the convex SimT can implicitly push the learned feature distribution of open-set classes away from closed-set classes.

We summarize our contributions in three aspects. 
\begin{itemize}
 \item We present a general and practical DA semantic segmentation framework to robustly learn from noisy pseudo-labeled target data, including closed-set and open-set label noises. To our best knowledge, it represents the first attempt to address the problem of open-set noisy label in DA.

 \item We, for the first time, model the closed-set and open-set noise distributions of target pseudo labels through SimT, and utilize computational geometry analysis to estimate it. The optimized SimT is used to benefit loss correction for noisy pseudo-labeled target data.

 \item Extensive experiments are conducted to verify the effect of the proposed SimT and demonstrate that it can be plugged into existing DA methods, such as UDA or SFDA, to boost the performance.
 \end{itemize}

\section{Related Work}
\subsection{Domain Adaptive Semantic Segmentation}
DA semantic segmentation aims to adapt a model for the target domain through transferring knowledge from a source domain. Unsupervised domain adaptation (UDA) methods leverage adversarial learning to align data distributions of two domains  \cite{tsai2018learning, kim2020learning, cheng2021dual}. Source-free domain adaptation (SFDA) methods \cite{kundu2021generalize, liu2021source, zhao2021source, prabhu2021s4t} mine useful knowledge embedded in source model to perform DA without source data. These DA works require the coexistence of source data (or the detailed specification of source models) and target data since concurrent access can help to characterize the domain gap \cite{kundu2021generalize}. However, it is not always practical in real-world deployment scenarios, where the source domain data transmission may be inefficient when it is extremely large \cite{zhao2021source}, data sharing may be restrained due to privacy \cite{prabhu2021s4t} or the specific knowledge of source models may be not accessible \cite{liang2021distill}. Different from UDA and SFDA, we study a more general and practical DA problem where only pseudo-labeled target data is accessible through a black-box model. 

Without specific requirements of source domain data or model, self-training strategy is already a common choice to adapt the model for unlabeled target data \cite{iqbal2020mlsl, pan2020unsupervised, zheng2021rectifying, zou2018unsupervised, liu2021source, zhang2021prototypical, guo2021metacorrection}. It optimizes the segmentation model with pseudo-labeled target data, which enhances the discriminative feature representation of target data \cite{guo2021metacorrection}. To tackle the target pseudo label noise issue, some methods filter out noisy pixels with respect to confidence or uncertainty \cite{iqbal2020mlsl, pan2020unsupervised, zheng2021rectifying, liu2021source}. Kundu \textit{et al.} \cite{liu2021source} train the source model with virtually extended multi-source data and select top 33\% confident target pixels for self-training. Zheng \textit{et al.} \cite{zheng2021rectifying} utilize uncertainty estimation to enable dynamic threshold. Instead of filtering out potentially noisy pseudo-labeled target data, other methods denoise pseudo labels \cite{zhang2021prototypical} or correct supervision signal \cite{guo2021metacorrection}. Zhang \textit{et al.} \cite{zhang2021prototypical} learn segmentation net from denoised pseudo labels according to class prototypes. Guo \textit{et al.} \cite{guo2021metacorrection} model the noise distribution of pseudo labels by NTM and solve it in a meta-learning strategy.

However, almost all self-training based methods ignore the open-set pseudo label noises in target domain, which are commonly encountered in practice \cite{liu2021exploiting}. Although confidence or uncertainty based self-training methods omit potentially noisy data and may reject some open-set pseudo label noises, simply dropping pixels will lead to a biased data distribution learning especially when pixels lie around the decision boundary. In this paper, we model the closed-set and open-set noise distributions of pseudo labels in target domain with a SimT, and optimize the segmentation model with corrected supervision signals on all target data. 

\subsection{Deep Learning with Noisy Labels}
Types of noisy labels can be divided into two categories: 

\noindent \textbf{Closed-set noisy labels.} `Closed-set' means that true class labels of noisy labeled samples are within known classes of noisy data. Many efforts have been devoted to tackling the closed-set noisy label issue, such as \textit{sample reweighting} \cite{shu2019meta}, \textit{label correction} \cite{zhang2020distilling}, \textit{loss correction} \cite{wang2020training, guo2021metacorrection}. Shu \textit{et al.} \cite{shu2019meta} leverage multiple layer perceptions to automatically suppress noisy samples. Zhang \textit{et al.} \cite{zhang2020distilling} conduct a dynamic linear combination of noisy label and prediction to refurbish labels. Wang \textit{et al.} \cite{wang2020training} utilize a set of trusted clean-labeled samples to estimate NTM for loss correction. 

\noindent \textbf{Open-set noisy labels.} `Open-set' indicates that true class labels of noisy labeled samples are outside known classes of noisy data. Contrast to the widely explored closed-set noisy label problem, it has been seldom invested, which is our focus in this study. Prior works concentrate on \textit{resampling} approach \cite{wang2018iterative, sachdeva2021evidentialmix, liu2021exploiting}. This solution is intuitive but suboptimal as the discarded data may contain useful information for shaping the decision boundary. \textit{Label correction} and \textit{loss correction} are vulnerable with open-set label noises since these methods are based on the assumption that training samples belong to known classes in the dataset. Hence, Yu \textit{et al.} \cite{yu2020unknown} transfer the `open-set' problem to `closed-set' one and then correct noisy labels. Xia \textit{et al.} \cite{xia2020extended} extend the traditional NTM with an open-set class, so as to model the open-set label noises. Our work represents the first effort to combat open-set noisy labels in DA.

\section{Methodology}

\begin{figure*}[htp!]
\centering
\includegraphics[width=173mm]{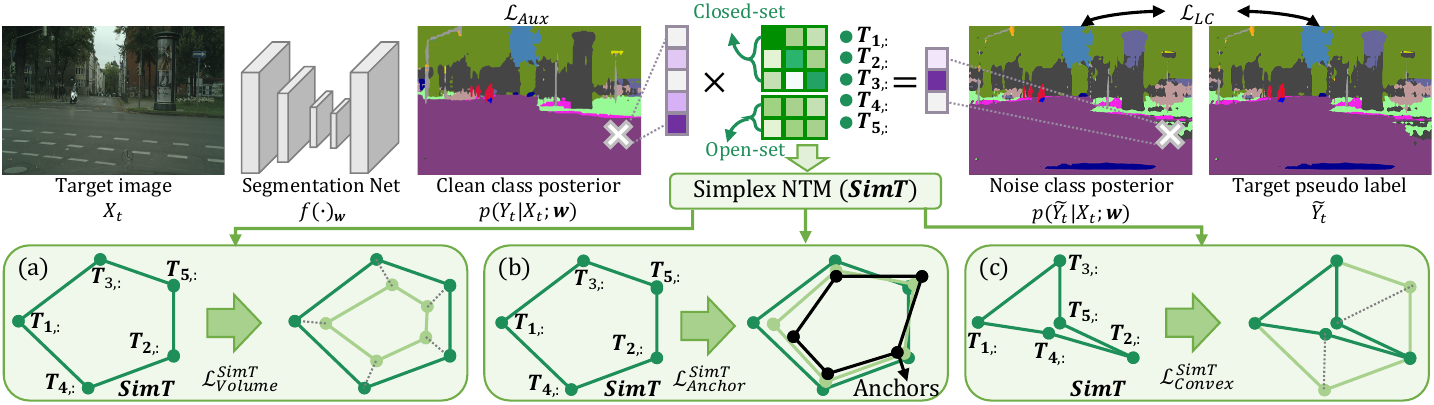}
\caption{The proposed domain adaption framework contains a segmentation net and a SimT. Target images are passed through the segmentation net $f(\cdot)_\mathbf{w}$ to perform semantic segmentation, and supervision signals are corrected by the learnable SimT. SimT is estimated through the proposed (a) volume regularization, (b) anchor guidance and (c) convex guarantee.}
\label{fig:Framework}
\end{figure*}

\subsection{Self-training with SimT} 
We focus on the problem of domain adaptive semantic segmentation, wherein we only have access to target domain images $\mathcal{X}_{\mathcal{T}} = \{X_{t} \in \mathbb{R}^{H\times W\times 3}\}_{t\in \mathcal{T}}$ with spatial dimension of $H\times W$ and the corresponding pseudo labels $\widetilde{\mathcal{Y}}_{\mathcal{T}} =\{\widetilde{Y}_{t}\in \{0, 1\}^{H\times W\times C} \}_{t\in \mathcal{T}}$. Note that $\widetilde{Y}_{t}$ is the pixel-wise label map with $C$ classes, and pseudo labels are derived from a given black-box model $f_{b}(\cdot)$. 
We aim to learn a segmentation model $f(\cdot)_\mathbf{w}$ parameterized by $\mathbf{w}$ that can correctly categorize pixels for target data $\mathcal{X}_{\mathcal{T}}$. Self-training based methods \cite{zou2018unsupervised, zheng2021rectifying, zhang2021prototypical, guo2021metacorrection} utilize pseudo labels as approximated ground truth labels for model training, and the cross-entropy loss over the target dataset for self-training is $\mathcal{L}_{ST}=-\sum_{t\in \mathcal{T}} \widetilde{Y}_{t} \log f\left({X}_{t}\right)_\mathbf{w}.$ 
By minimizing $\mathcal{L}_{ST}$ on target data, the optimized model can be discriminatively adapted to the target domain. However, noises in pseudo-labeled target domain data, including closed-set label noises and open-set label noises, inevitably deteriorate the performance of self-training based DA methods.

To enhance the noise tolerance property of $\mathcal{L}_{ST}$, we propose SimT to benefit loss correction, as shown in Figure \ref{fig:Framework}. SimT is formulated as $\boldsymbol{T} \in [0, 1]^{(C+n)\times C}$, where $\boldsymbol{T}_{1:C,:}$ models closed-set noise distribution and $\boldsymbol{T}_{C+1:C+n,:}$ models open-set noise distribution in target domain. $n$ is a hyper-parameter that indicates the potential open-set class number, and we set $n>1$ to implicitly model the diverse semantics within open-set classes. The defined SimT specifies the probability of ground truth label (${Y}_{t}=j$) flipping to noisy label ($\widetilde{Y}_{t}=k$) by $\boldsymbol{T}_{jk} = p(\widetilde{Y}_{t} = k\mid {Y}_{t} = j)$, and we have $\sum_{k=1}^{C} \boldsymbol{T}_{jk}=1$. Assuming the class posterior probability for noisy pseudo label is $p(\widetilde{Y}_{t} = k \mid X_{t}; \mathbf{w}) \in [0, 1]^{C}$ and for ground truth label is $p({Y}_{t}=j \mid X_{t}; \mathbf{w}) \in [0, 1]^{C+n}$, SimT bridges the generated pseudo label $\widetilde{{Y}}_{t}$ and the ground truth label ${{Y}_{t}}$ through: 
\begin{equation}
\begin{aligned}
\begin{split}
p(\widetilde{Y}_{t} = k \mid X_{t}; \mathbf{w})&=\sum_{j=1}^{C+n} \boldsymbol{T}_{jk} \cdot p({Y}_{t}=j \mid X_{t}; \mathbf{w}),\\
\Rightarrow p(\widetilde{Y}_{t} \mid X_{t}; &\mathbf{w})= p({Y}_{t} \mid X_{t}; \mathbf{w})~ \boldsymbol{T}.
\label{SimT}
\end{split}
\end{aligned}
\end{equation}
With the formulated noise distribution ($\boldsymbol{T}$), we correct the self-training loss over noisy pseudo-labeled target data as 
\begin{equation}
\mathcal{L}_{LC}=-\sum_{t\in \mathcal{T}} \widetilde{Y}_{t} \log [f\left({X}_{t}\right)_\mathbf{w} \boldsymbol{T}]. 
\label{LC}
\end{equation}
This corrected loss encourages the similarity between noisy class posterior probabilities $p(\widetilde{Y}_{t} \mid X_{t}; \mathbf{w}) = f\left({X}_{t}\right)_\mathbf{w} \boldsymbol{T}$ and noisy pseudo labels $\widetilde{Y}_{t}$. It is obvious that once the true SimT is obtained, we can recover the desired estimation of clean class posterior probability $p({Y}_{t}|X_{t}; \mathbf{w})$ by the softmax of first $C$ classes output $f\left({X}_{t}\right)_\mathbf{w}$ even training the segmentation model with noisy data. Considering no clean-labeled target domain data is accessible, previous work estimates NTM through a meta-learning strategy \cite{guo2021metacorrection}. Different from prior art, we formulate NTM to make up a simplex and solve it through computational geometry analysis without the requirement of a clean labeled target data set. 

\subsection{Volume Regularization of SimT}\label{volume}
Semantic segmentation can be regarded as pixel-wise classification, thus we make geometry analysis at pixel level. Given a pixel $\boldsymbol{x}$ within target domain image $X_{t}$, the noisy class posterior probability $p(\widetilde{\boldsymbol{y}} \mid \boldsymbol{x})=[p(\widetilde{y}=1 \mid \boldsymbol{x}), ~p(\widetilde{y}=2 \mid \boldsymbol{x}),~ ..., ~p(\widetilde{y}=C \mid \boldsymbol{x})]$ can be regarded as a point in $C$-dimensional space, and the clean class posterior probability $p(\boldsymbol{y} \mid \boldsymbol{x})=[p({y}=1 \mid \boldsymbol{x}), ~p({y}=2 \mid \boldsymbol{x}),~ ..., ~p({y}=C+n \mid \boldsymbol{x})]$ is a $(C+n)$ dimensional vector. 
These two class posteriors can be bridged by the proposed SimT via $p(\widetilde{\boldsymbol{y}} \mid \boldsymbol{x})= p({\boldsymbol{y}} \mid \boldsymbol{x}) \boldsymbol{T}$ as in Eq. (\ref{SimT}). In the meanwhile, conditions of $\sum_{j=1}^{C+n} p({y}=j \mid \boldsymbol{x})=1$ and $p({\boldsymbol{y}} \mid \boldsymbol{x})>0$ are satisfied, thus, the noisy class posterior probability $p(\widetilde{\boldsymbol{y}} \mid \boldsymbol{x})$ can be interpreted as a convex combination of $\boldsymbol{T}$ rows with factor of $p({\boldsymbol{y}} \mid \boldsymbol{x})$. In other words, the simplex formed by rows of $\boldsymbol{T}$ encloses $p(\widetilde{\boldsymbol{y}} \mid \boldsymbol{x})$ for any input pixel $\boldsymbol{x}$ \cite{li2021provably}. However, SimT is not identifiable if no assumption is made since there exists an infinite number of simplices enclosing $p(\widetilde{\boldsymbol{y}} \mid \boldsymbol{x})$. 

To approximate the true SimT, we follow the sufficiently scattered assumption \cite{li2021provably} and devise volume regularization. Among all simplices enclosing $p(\widetilde{\boldsymbol{y}} \mid \boldsymbol{x})$, the true SimT should be the one with minimum volume under the sufficiently scattered assumption. In this circumstance, the volume consisted by the target clean class posterior will converge the maximum one, and outputs of optimized segmentation model thus approximate to the ground truth label distribution. This motivates us to propose the following volume regularization for the optimization of SimT: 
\begin{equation}
\mathcal{L}_{Volume}^{SimT} = \mathrm{log} [ \mathrm{Vol}( \boldsymbol{T})] = \mathrm{log} \sqrt{\mathrm{det}(\boldsymbol{T}^{\top}\boldsymbol{T})} / (N !)
\label{Volume}
\end{equation}
where $\sqrt{\operatorname{det}(\boldsymbol{T}^{\top} \boldsymbol{T})} /(N !)$ denotes the volume of a r-parallelogram defined by the non-square matrix $\boldsymbol{T}$ in geometry \cite{gritzmann1995largest, fu2016robust}. Since $N!$ is a constant number, we discard this denominator in experiments. $\mathrm{log}(\cdot)$ function is to stabilize the optimization for computed determinant, following \cite{fu2016robust}.

To make $\boldsymbol{T}$ differentiable and satisfy the condition of $\boldsymbol{T} \in [0, 1]^{(C+n)\times C}$, $\sum_{k=1}^{C} \boldsymbol{T}_{jk}=1$, we utilize the reparameterization method. Specifically, we first randomly initialize a matrix $\boldsymbol{U} \in \mathbb{R}^{(C+n)\times C}$. To preserve the diagonally dominant property of closed-set part of SimT ($\boldsymbol{T}_{1:C,:}$), the diagonal prior $\boldsymbol{I}$ is introduced, which has been widely used in the literature of NTM estimation \cite{patrini2017making, li2021provably}. Considering segmentation model tends to classify samples into majority categories, the class distribution of pseudo labels $\boldsymbol{C}$ is involved. Incorporating these prior informations, we obtain $\boldsymbol{V} = \boldsymbol{C} \cdot \sigma (\boldsymbol{U}) + \boldsymbol{I}$, where $\sigma(\cdot)$ is the sigmoid function to avoid negative value in SimT. Then we do the normalization of $\boldsymbol{T}_{jk} = \frac{\boldsymbol{V}_{jk}}{\sum_{k=1}^{C} \boldsymbol{V}_{jk}}$ to derive SimT ($\boldsymbol{T}$). Since both sigmoid function and normalization operation are differentiable, SimT can be updated through gradient descent on $\boldsymbol{U}$. 

\subsection{Anchor Guidance of SimT}\label{anchor}
Considering the diagonally dominant assumption is not available for the open-set part $\boldsymbol{T}_{C:C+n,:}$ of SimT, the open-set part will eventually collapse to a trivial solution by minimizing the volume regularization $\mathcal{L}_{Volume}^{SimT}$. To prevent the trivial solution of open-set part in SimT, we devise anchor guidance, where anchor points are detected to guide the estimation of SimT. Formally, anchor point of class $c$ is defined as the pixel whose clean class posterior are $p(y=c \mid \boldsymbol{x}^{c})=1$ and $p(y \neq c \mid \boldsymbol{x}^{c})=0$. Thus we have $p(\widetilde{y}=k \mid \boldsymbol{x}^{c}) = \sum_{j=1}^{C+n} \boldsymbol{T}_{jk} \cdot p(y=j \mid \boldsymbol{x}^{c}) = \boldsymbol{T}_{ck}$, indicating that the noisy class posterior probability of anchor point $p(\widetilde{\boldsymbol{y}} \mid \boldsymbol{x}^{c})$ lies at one vertex of SimT ($\boldsymbol{T}_{c,:}$). Hence, the simplex formed by noisy class posteriors of anchor points can recover the true SimT in theory. 
However, under mixed open-set and closed-set noisy labels, anchor points are hard to detect. Therefore, we define anchor point as the most representative pixel within each closed-set and open-set class with the largest clean class posterior probability. The anchor point in class $c$ is thus derived via 
\begin{equation}
\boldsymbol{x}^{c} = \underset{\boldsymbol{x}}{\mathrm{argmax}} ~ p(y=c \mid \boldsymbol{x}; \mathbf{w}).
\label{Anchor}
\end{equation}
In geometrical perspective, the estimated $\boldsymbol{x}^{c}$ is the point, whose noisy class posterior $p(\widetilde{\boldsymbol{y}} \mid \boldsymbol{x}^{c})$ is closest to the vertex $\boldsymbol{T}_{c,:}$ in $C$-dimensional space. Then we leverage computed anchor points to guide the approximation of the true SimT, and the anchor guidance is formulated as 

\begin{small}
\begin{equation}
\mathcal{L}_{Anchor}^{SimT} = \sum_{c=1}^{C+n} \sum_{k=1}^{C} \left \| \boldsymbol{T}_{ck} - p(\widetilde{y}=k \mid \boldsymbol{x}^{c}; \mathbf{w}^{fixed})  \right \|^{2},
\label{Anchor}
\end{equation}
\end{small}where the noisy class posterior of anchor point $p(\widetilde{y}=k \mid \boldsymbol{x}^{c}; \mathbf{w}^{fixed})$ is computed from the segmentation model warmed up with pseudo labels, denoted as $f(\cdot)_{\mathbf{w}^{fixed}}$ with parameter of $\mathbf{w}^{fixed}$. For each target image, only occurred classes would derive the corresponding anchor points. Then those detected anchor points are leveraged to calculate the regularization of anchor guidance. The accurate detection of anchor points in open-set classes can prevent the trivial solution and guide to approximate the true SimT.

\textbf{Auxiliary loss for anchor point detection.} 
To facilitate the detection of anchor points in open-set classes, an auxiliary loss enhancing the discriminative feature learning among closed-set and open-set data is designed. Firstly, we introduce a threshold to select confident pixels as non-noisy samples, and class posterior probabilities of those selected pixels are directly supervised to optimize the segmentation model. This direct supervision explicitly enables the model to encode the diverse semantics for both closed-set and open-set classes. Specifically, the confidence score is derived from the segmentation model $f(\cdot)_{\mathbf{w}^{fixed}}$. We select confident closed-set pixels $X_{\mathcal{K}}=\{\boldsymbol{x}_{k} \}_{k\in \mathcal{K}}$ through the condition of $\mathrm{max} ~f(\boldsymbol{x}_{k})_{\mathbf{w}^{fixed}} > \tau_{high}$, and the corresponding labels $\widetilde{\boldsymbol{y}}_{k}$ are one-hot vectors of $\widetilde{y}_{k}=\mathrm{argmax}~ f(\boldsymbol{x}_{k})_{\mathbf{w}^{fixed}}$. Confident open-set pixels $X_{\mathcal{U}}=\{\boldsymbol{x}_{u} \}_{u\in \mathcal{U}}$ are those with $\mathrm{max} ~f(\boldsymbol{x}_{u})_{\mathbf{w}^{fixed}} < \tau_{low} ~\& ~\mathrm{argmax}~ f(\boldsymbol{x}_{u})_{\mathbf{w}} > C$, and their one-hot ground truth labels $\widetilde{\boldsymbol{y}}_{u}$ are derived from $\widetilde{y}_{u}=\mathrm{argmax}~ f(\boldsymbol{x}_{u})_{\mathbf{w}}$. Note that $\tau_{high}$, $\tau_{low}$ are hyper-parameters and empirically set to be 0.8, 0.2. To further detect open-set class samples, we make the open-set classifier to output the second-largest probability for closed-set pixels $X_{\mathcal{K}}$, which is constrained by the second term of following auxiliary loss:
\begin{equation}
\begin{aligned}
\begin{split}
\mathcal{L}_{Aux}=&- \sum_{i\in \mathcal{K} \& \mathcal{U}} \widetilde{\boldsymbol{y}}_{i} \log f\left(\boldsymbol{x}_{i} \right)_{\mathbf{w}} \\
&- \lambda \sum_{k\in \mathcal{K}}\widetilde{\boldsymbol{y}}^{o}\log [f\left(\boldsymbol{x}_{k}\right)_{\mathbf{w}} \setminus \widetilde{y}_{k}] , 
\label{Aux}
\end{split}
\end{aligned}
\end{equation}
where $\widetilde{\boldsymbol{y}}^{o}$ is the one-hot form of $\widetilde{y}^{o}=\underset{c\in [C, C+n]}{\mathrm{argmax}}~f(\boldsymbol{x}_{k})_{\mathbf{w}}$, and $\lambda$ is a hyper-parameter to control the contribution of the second term in auxiliary loss. The first item maintains performance in the closed-set classes and detects open-set class pixels. The second term matches the masked-probability with the most possible open-set class, which prevents the open-set classifier far away from closed-set classifier and facilitates the discovery of open-set class pixels. The discriminative feature learning induced by auxiliary loss simplifies the detection of anchor points, hence enhancing the performance of anchor guidance.

\begin{table*}[tp!]
\centering
\vspace{+0.2cm}
\caption{Results of UDA and SFDA on adapting GTA5 to Cityscapes.}
\label{UDA_SFDA}
\scalebox{0.69}{\begin{tabular}{c | c  c  c  c  c  c  c  c  c  c  c  c  c  c  c  c  c  c  c | c}
\ChangeRT{1.5pt}
Methods&\rotatebox{90}{road}&\rotatebox{90}{sidewalk}&\rotatebox{90}{building}&\rotatebox{90}{wall}&\rotatebox{90}{fence}&\rotatebox{90}{pole}&\rotatebox{90}{t-light}&\rotatebox{90}{t-sign}&\rotatebox{90}{veg.}&\rotatebox{90}{terrain}&\rotatebox{90}{sky}&\rotatebox{90}{person}&\rotatebox{90}{rider}&\rotatebox{90}{car}&\rotatebox{90}{truck}&\rotatebox{90}{bus}&\rotatebox{90}{train}&\rotatebox{90}{motor}&\rotatebox{90}{bike}&{mIoU}\\

\ChangeRT{1.3pt}
\multicolumn{21}{c}{UDA}\\
\ChangeRT{1.pt}







AdaptSegNet (18') \cite{tsai2018learning}&86.5&36.0&79.9&23.4&23.3&23.9&35.2&14.8&83.4&33.3&75.6&58.5&27.6&73.7&32.5&35.4&3.9&30.1&28.1&42.4 \\

LTIR (20') \cite{kim2020learning}&{92.9}&55.0&85.3&34.2&31.1&34.9&40.7&34.0&85.2&{40.1}&{87.1}&61.0&31.1&82.5&32.3&42.9&0.3&36.4&46.1&50.2 \\

MLSL (20') \cite{iqbal2020mlsl}&89.0&45.2&78.2&22.9&27.3&{37.4}&{46.1}&{43.8}&82.9&18.6&61.2&60.4&26.7&85.4&35.9&44.9&\textbf{36.4}&{37.2}&{49.3}&49.0 \\

IntraDA (20') \cite{pan2020unsupervised}&90.6&37.1&82.6&30.1&19.1&29.5&32.4&20.6&{85.7}&{40.5}&79.7&58.7&31.1&{86.3}&31.5&{48.3}&0.0&30.2&35.8&46.3 \\

RPLL (21') \cite{zheng2021rectifying}&90.4&31.2&85.1&{36.9}&25.6&{37.5}&{48.8}&{48.5}&85.3&34.8&81.1&{64.4}&{36.8}&{86.3}&34.9&{52.2}&1.7&29.0&44.6&50.3 \\

MetaCorrection (21') \cite{guo2021metacorrection}&{92.8}&{58.1}&{86.2}&{39.7}&{33.1}&36.3&{42.0}&38.6&85.5&37.8&{87.6}&{62.8}&31.7&84.8&35.7&{50.3}&2.0&{36.8}&{48.0}&{52.1} \\

DPL-Dual (21') \cite{cheng2021dual}&92.8&54.4&86.2&\underline{41.6}&32.7&36.4&\underline{49.0}&34.0&85.8&41.3&86.0&63.2&34.2&87.2&39.3&44.5&\underline{18.7}&42.6&43.1&53.3 \\

UncerDA (21') \cite{wang2021uncertainty}&90.5&38.7&86.5&41.1&32.9&40.5&48.2&42.1&86.5&36.8&84.2&64.5&38.1&87.2&34.8&50.4&0.2&41.8&54.6&52.6 \\

DSP (21') \cite{gao2021dsp}&92.4&48.0&87.4&33.4&35.1&36.4&41.6&46.0&87.7&43.2&\textbf{89.8}&66.6&32.1&{89.9}&\underline{57.0}&56.1&0.0&44.1&\underline{57.8}&55.0 \\

ProDA (21') \cite{zhang2021prototypical}&87.8&56.0&79.7&\textbf{46.3}&\textbf{44.8}&\textbf{45.6}&\textbf{53.5}&53.5&\textbf{88.6}&45.2&82.1&70.7&39.2&88.8&45.5&\underline{59.4}&1.0&\underline{48.9}&56.4&57.5 \\

BAPA-Net (21') \cite{liu2021bapa}&\textbf{94.4}&\textbf{61.0}&\underline{88.0}&26.8&\underline{39.9}&38.3&46.1&\underline{55.3}&87.8&\underline{46.1}&\underline{89.4}&\underline{68.8}&\underline{40.0}&\underline{90.2}&\textbf{60.4}&59.0&0.0&45.1&54.2&57.4 \\

\textbf{Ours (SimT)}&\underline{94.2}&\underline{60.0}&\textbf{88.5}&{30.3}&{39.7}&\underline{41.2}&47.8&\textbf{60.8}&\textbf{88.6}&\textbf{47.3}&89.3&\textbf{71.5}&\textbf{45.0}&\textbf{90.7}&54.2&\textbf{60.2}&0.0&\textbf{51.8}&\textbf{58.4}&\textbf{58.9} \\

\ChangeRT{1.3pt}
\multicolumn{21}{c}{SFDA}\\
\ChangeRT{1.pt}
Source only&75.8&16.8&77.2&12.5&21.0&25.5&30.1&20.1&81.3&24.6&70.3&53.8&26.4&49.9&17.2&25.9&6.5&25.3&36.0&36.6\\

Test-time BN (20') \cite{nado2020evaluating}&79.5&79.5&81.3&72.7&29.8&74.1&28.0&58.8&25.3&22.2&19.4&11.0&22.6&17.0&24.5&19.9&\underline{34.3}&2.4&14.7&37.7\\

SHOT (20') \cite{liang2020shot}&88.7&34.9&82.1&27.6&22.5&30.9&31.4&24.7&83.6&37.7&76.8&58.1&24.9&83.9&34.1&39.5&6.0&26.8&24.6&44.1\\

TENT (21') \cite{wang2021tent}&87.3&79.8&83.8&\textbf{85.0}&39.0&77.7&21.2&57.9&34.7&19.6&24.3&4.5&16.6&20.8&24.9&17.8&25.1&2.0&16.6&38.9\\

SFDA (21') \cite{liu2021source}&84.2&\underline{82.7}&82.4&80.0&\underline{39.2}&\textbf{85.3}&25.9&\underline{58.7}&30.5&22.1&27.5&30.6&21.9&{31.5}&33.1&22.1&31.1&3.6&27.8&43.2\\

S4T (21') \cite{prabhu2021s4t}&89.7&\textbf{84.4}&\textbf{86.8}&\textbf{85.0}&\textbf{46.7}&\underline{79.3}&\underline{39.5}&\textbf{61.2}&41.8&29.0&25.7&9.3&\textbf{36.8}&28.2&19.3&26.7&\textbf{45.1}&5.3&11.8&44.8\\

URMA (21') \cite{prabhu2021uncertainty}&92.3&55.2&81.6&30.8&18.8&37.1&17.7&12.1&84.2&35.9&83.8&57.7&24.1&81.7&27.5&44.3&6.9&24.1&40.4&45.1\\

SFDASeg (21') \cite{kundu2021generalize}&\underline{91.7}&53.4&86.1&37.6&32.1&37.4&38.2&35.6&\underline{86.7}&\textbf{48.5}&\textbf{89.9}&\underline{62.6}&34.3&\underline{87.2}&\textbf{51.0}&\underline{50.8}&4.2&\underline{42.7}&\underline{53.9}&\underline{53.4} \\

\textbf{Ours (SimT)}&\textbf{92.3}&55.8&\underline{86.3}&34.4&31.7&37.8&\textbf{39.9}&41.4&\textbf{87.1}&\underline{47.8}&\underline{88.5}&\textbf{64.7}&\underline{36.3}&\textbf{87.3}&\underline{41.7}&\textbf{55.2}&0.0&\textbf{47.4}&\textbf{57.6}&\textbf{54.4} \\
\ChangeRT{1.5pt}
\end{tabular}}
\end{table*}

\subsection{Convex Guarantee of SimT}\label{convex}
As analyzed in \S \ref{sec:intro}, the trivial solution of open-set part will lead to the non-convex property of SimT. In other word, one of vertices in SimT may be the convex combination of the rest vertices. This means different ground truth classes may show the similar noise transition probability, and results in multiple clean class posteriors $p(\boldsymbol{y} \mid \boldsymbol{x})$ correspond to the same noisy class posterior $p(\widetilde{\boldsymbol{y}} \mid \boldsymbol{x})$. Considering similar noise transition probabilities reveal close feature representations \cite{xia2020extended}, the non-convex SimT may lead the segmentation model to produce undiscriminating features. Hence, we advance convex guarantee to prevent SimT from becoming non-convex. 
Inspired by the insight that each vertex of the convex hull cannot be represented by any convex combinations of rest vertices, we first introduce a weighting matrix $\boldsymbol{u} \in [0, 1]^{(C+n)\times (C+n)}$ with constraints of $\boldsymbol{u}_{j,k=j} = -1$ and $\sum_{k}\boldsymbol{u}_{j,k\neq j} =1$ to denote convex combination coefficients. The reparameterization method is introduced to make the weighting matrix $\boldsymbol{u}$ differentiable and satisfy its constraints. 
Then the convex guarantee is formulated as a minimax problem, where the optimization objective is $\underset{\boldsymbol{T}}{\mathrm{max}}~\underset{\boldsymbol{u}}{\mathrm{min}} \left \|  \boldsymbol{u}\boldsymbol{T} \right \|^{2}$.
By updating $\boldsymbol{u}$ to minimize the optimization objective, the weighting matrix learns to represent each vertex with the convex combination of rest vertices. In turn, by updating $\boldsymbol{T}$ to maximize the optimization objective, each vertex is pushed away from the convex combination of rest vertices, and SimT is encouraged to be convex. For simplification, convex guarantee regularization is formulated as
\begin{equation}
\mathcal{L}_{Convex}^{SimT} = -  \left \|  \boldsymbol{u}\boldsymbol{T} \right \|^{2}, ~\mathrm{where}~ \boldsymbol{u} = \underset{\boldsymbol{u}}{\mathrm{argmin}} \left \|  \boldsymbol{u}\boldsymbol{T} \right \|^{2}. 
\label{Convex}
\end{equation}
The convex guaranteed SimT enforces the class posterior distribution to be scattered, thereby the segmentation model is encouraged to extract discriminative features among closed-set and open-set classes.

\subsection{Optimization summary for SimT} With the proposed regularizers, SimT is optimized to estimate the mixed closed-set and open-set noise distributions, thereby benefiting the loss correction of pseudo-labeled target data. In summary, the overall objective function for our DA semantic segmentation framework is formulated as \begin{equation}
\begin{aligned}
\begin{split}
\mathcal{L}(\mathbf{w}, \boldsymbol{T}) = &\mathcal{L}_{LC}(\mathbf{w}, \boldsymbol{T}) + \mathcal{L}_{Aux}(\mathbf{w}) + \\
\alpha \mathcal{L}_{Volume}^{SimT} &(\boldsymbol{T}) +
\beta \mathcal{L}_{Anchor}^{SimT}(\boldsymbol{T}) + \gamma \mathcal{L}_{Convex}^{SimT}(\boldsymbol{T}).
\label{Total}
\end{split}
\end{aligned}
\end{equation}
where corrected loss $\mathcal{L}_{LC}(\mathbf{w}, \boldsymbol{T})$ (Eq. (\ref{LC})) and $\mathcal{L}_{Aux}(\mathbf{w})$ (Eq. (\ref{Aux})) optimize the segmentation model. $\mathcal{L}_{LC}(\mathbf{w}, \boldsymbol{T})$ and three regularizers $\mathcal{L}_{Volume}^{SimT}(\boldsymbol{T})$ (Eq. (\ref{Volume})), $\mathcal{L}_{Anchor}^{SimT} (\boldsymbol{T})$ (Eq. (\ref{Anchor})), $\mathcal{L}_{Convex}^{SimT}(\boldsymbol{T})$ (Eq. (\ref{Convex})) are minimized to approximate the true SimT. $\alpha$, $\beta$, $\gamma$ are regularization coefficients to balance contributions of three regularizers.

\section{Experiments}
\subsection{Training Details}
\textbf{Datasets.} 
We evaluate the performance of our method on two DA semantic segmentation tasks, including a synthetic-to-real scenario (GTA5 \cite{richter2016playing}$\rightarrow$Cityscapes \cite{cordts2016cityscapes}) and a surgical instrument type segmentation task (Endovis17 \cite{allan20192017}$\rightarrow$ Endovis18 \cite{allan20202018}). \textbf{\textit{GTA5}} contains 24,966 images captured from a video game, and 19 classes in pixel-wise annotations are compatible with Cityscapes. \textbf{\textit{Cityscapes}} is a real-world semantic segmentation dataset obtained in driving scenarios, containing 34 classes according to the manual annotations. Specifically, 2,975 unlabeled images are regarded as the target domain training data, and evaluations are performed on 500 validation images with manual annotations. 
\textbf{\textit{Endovis17}} contains 1800 source images captured from abdominal porcine procedure, and 3 instrument type classes are compatible with Endovis18. \textbf{\textit{Endovis18}} is an abdominal procedure dataset, consisting of 2384 target images with 6 instrument types. Following \cite{liu2021ignet}, we divide Endovis18 into 1639 unlabeled training images and 596 validation images.

\textbf{Implementation Details.} 
We adopt DeepLab-v2 \cite{chen2017deeplab} backbone with encoder of ResNet-101 \cite{he2016deep} as our segmentation model. Given the pseudo-labeled target domain data derived from a black-box model $f_{b}(\cdot)$, we warm up the segmentation model to obtain $f(\cdot)_{\mathbf{w}^{fixed}}$ with $C$-way output probabilities. The classifier of segmentation model is then extended to output $(C+n)$-way clean class posterior probabilities. The extended model is denoted as $f(\cdot)_{\mathbf{w}}$ we aim to optimize, and we obtain final predictions from first $C$-way of $f(\cdot)_{\mathbf{w}}$ during inference phase. Our method is implemented with the PyTorch library on Nvidia 2080Ti. We adopt polynomial learning rate scheduling to optimize the feature extractor with the initial learning rate of 6e-4, while it is set as 6e-3 for the optimization of classifier and SimT. The batch size is set as 1, and the maximum iteration number is 40000. Hyper-parameters of $n$, $\lambda$, $\alpha$, $\beta$, $\gamma$ are set as 15, 0.1, 1.0, 1.0, 0.1 in our implementation. Performances are evaluated by the widely utilized metrics, intersection-over-union (IoU) of each class and the mean IoU (mIoU).

\subsection{Results on GTA5$\rightarrow$Cityscapes}

\begin{table}[tp!]
\vspace{+0.2cm}
\centering
\caption{Results on adapting Endovis17 to Endovis18.}
\label{Endovis17_18}
\scalebox{0.85}{\begin{tabular}{c | c  c  c  | c}
\ChangeRT{1.5pt}
Methods&scissor&needle driver&forceps&{mIoU}\\

\ChangeRT{1.3pt}
\multicolumn{5}{c}{UDA}\\
\ChangeRT{1.pt}

I2I (19') \cite{pfeiffer2019generating}&60.78&15.34&52.55&42.89\\

FDA (20') \cite{yang2020fda}&68.25&18.29&50.29&45.61\\


S2RC (20') \cite{sahu2020endo}&64.48&22.79&49.34&45.54\\


IntraDA (20') \cite{pan2020unsupervised}&68.07&25.88&52.63&48.86\\

ASANet (20') \cite{zhou2020affinity}&70.19&11.34&50.70&44.08\\

EI-MUNIT(21') \cite{colleoni2021robotic}&62.73&29.83&48.32&46.96\\

RPLL (21') \cite{zheng2021rectifying}&65.11&13.17&51.75&43.34\\

IGNet (21') \cite{liu2021ignet}&\underline{70.65}&\underline{37.60}&\underline{52.97}&\underline{53.74}\\

\textbf{Ours (SimT)}&\textbf{76.24}&\textbf{39.83}&\textbf{58.94}&\textbf{58.34} \\

\ChangeRT{1.3pt}
\multicolumn{5}{c}{SFDA}\\
\ChangeRT{1.pt}

Source only&45.97&22.08&38.92&35.66\\


SHOT (20') \cite{liang2020shot}&\underline{68.44}&22.90&47.95&46.43 \\


SFDA (21') \cite{liu2021source}&66.05&24.90&45.79&45.58 \\

SFDASeg (21') \cite{kundu2021generalize}&67.90&\underline{35.31}&\underline{52.46}&\underline{51.89} \\

\textbf{Ours (SimT)}&\textbf{75.56}&\textbf{36.46}&\textbf{53.66}&\textbf{55.23} \\
\ChangeRT{1.5pt}
\end{tabular}}
\vspace{+0.2cm}
\end{table}

\textbf{UDA.} 
We first verify the effectiveness of our approach in the GTA5$\rightarrow$Cityscapes scenario, and the corresponding comparison results are listed in Table \ref{UDA_SFDA} with the first and second best results highlighted in bold and underline. Under the setting of UDA, our black-box model $f_{b}(\cdot)$ can be any existing UDA models, and we leverage model of \cite{liu2021bapa} to generate pseudo labels for unlabeled target domain data. Overall, the proposed SimT method arrives at  the state-of-the-art mIoU score of 58.9\%, surpassing all other UDA methods by a large margin. In terms of the per class IoU score, the proposed SimT method shows the superior performance. To be specific, we arrive at the best on 10 out of 19 categories and the second best on 3 out of 19 categories. Moreover, SimT outperforming other self-training based methods \cite{zou2018unsupervised, zou2019confidence, iqbal2020mlsl, pan2020unsupervised, zheng2021rectifying, guo2021metacorrection, gao2021dsp, zhang2021prototypical, liu2021bapa} demonstrates the effectiveness of the proposed method in mitigating the closed-set and open-set label noise problems.

\textbf{SFDA.} 
In the GTA5$\rightarrow$Cityscapes scenario, we then verify the proposed SimT method under the SFDA setting, where the black-box model $f_{b}(\cdot)$ is borrowed from a SFDA model in \cite{kundu2021generalize}. The corresponding comparison results of our method and baseline models are listed in Table \ref{UDA_SFDA}. It is observed that the proposed SimT surpasses all other SFDA methods with the state-of-the-art mIoU of 54.4\%, outperforming the model learned only from the source domain data by a large margin of 18.8\% in mIoU.

\subsection{Results on Endovis17$\rightarrow$Endovis18}

\textbf{UDA.} We further verify the effectiveness of SimT in a medical scenario, \ie Endovis17$\rightarrow$Endovis18, and the quantitative comparison results are listed in Table \ref{Endovis17_18}. We utilize the UDA model of \cite{liu2021ignet} to generate pseudo labels for unlabeled target domain data. Obviously, the proposed SimT method obtains the state-of-the-art performance with 58.12\% mIoU, overpassing all other UDA methods by a large margin. This observation proves the impact of the proposed method in medical image analysis.

\textbf{SFDA.} In the SFDA setting of Endovis17$\rightarrow$Endovis18, performance of the proposed SimT is evaluate based on that the black-box model $f_{b}(\cdot)$ is borrowed from \cite{kundu2021generalize}. The corresponding comparison results of our method and baseline models in Table \ref{Endovis17_18} show that the proposed SimT is superior to all other SFDA methods with mIoU of 55.23\%.

\subsection{Ablation Study}

\begin{table}[tp!] 
\vspace{+0.2cm}
\caption{Ablation study. `Pseudo Label' denotes we employ pseudo labels generated by the corresponding UDA model.}
\centering
\noindent
\scalebox{0.68}{\begin{tabular}{c | c | c  c }
\ChangeRT{1.5pt}
\xrowht{11pt}
Method&Pseudo Label&\begin{tabular}[c]{@{}c@{}}GTA5 $\rightarrow$ \\Cityscapes\end{tabular}&$\Delta$\\
\ChangeRT{1.0pt}
\xrowht{6pt}
AdaptSegNet (CVPR18') \cite{tsai2018learning}&---&42.4&--\\
\xrowht{6pt}
Self-training (MRENT, ICCV 19'\cite{zou2019confidence})&AdaptSegNet&45.1&2.7\\
\xrowht{6pt}
Self-training (Threshold, ECCV18' \cite{zou2018unsupervised})&AdaptSegNet&44.4&2.0\\
\xrowht{6pt}
Self-training (Ucertainty, IJCV21' \cite{zheng2021rectifying})&AdaptSegNet&46.1&3.7\\
\xrowht{6pt}
Self-training (MetaCorrection, CVPR21' \cite{guo2021metacorrection})&AdaptSegNet&47.3&4.9\\
\xrowht{6pt}
Ours (w/o $\mathcal{L}_{Volume}^{SimT}$)&AdaptSegNet&46.8&4.4\\
\xrowht{6pt}
Ours (w/o $\mathcal{L}_{Anchor}^{SimT}$)&AdaptSegNet&46.9&4.5\\
\xrowht{6pt}
Ours (w/o $\mathcal{L}_{Aux}$)&AdaptSegNet&46.7&4.3\\
\xrowht{6pt}
Ours (w/o $\mathcal{L}_{Convex}^{SimT}$)&AdaptSegNet&47.3&4.9\\
Ours&AdaptSegNet&\textbf{48.0}&\textbf{5.6}\\

\ChangeRT{1.0pt}
LTIR (CVPR20') \cite{kim2020learning}&---&50.2&--\\
Self-training (MetaCorrection, CVPR21' \cite{guo2021metacorrection})&LTIR&{52.1}&{1.9}\\
Ours&LTIR&\textbf{52.3}&\textbf{2.1}\\

\ChangeRT{1.0pt}
DSP (AAAI21'') \cite{gao2021dsp}&---&55.0&--\\
Self-training (MetaCorrection, CVPR21' \cite{guo2021metacorrection})&DSP&{56.4}&{1.4}\\
Ours&DSP&\textbf{57.3}&\textbf{2.3}\\

\ChangeRT{1.0pt}
BAPA-Net (ICCV21') \cite{liu2021bapa}&---&57.4&--\\
Self-training (MetaCorrection, CVPR21' \cite{guo2021metacorrection})&BAPA-Net&57.7&0.3\\
Ours&BAPA-Net&\textbf{58.9}&\textbf{1.5}\\

\ChangeRT{1.5pt}
\end{tabular}}
\label{ablation}
\vspace{+0.2cm}
\end{table}

\begin{figure}
\vspace{+0.2cm}
\centering
\includegraphics[width=84mm]{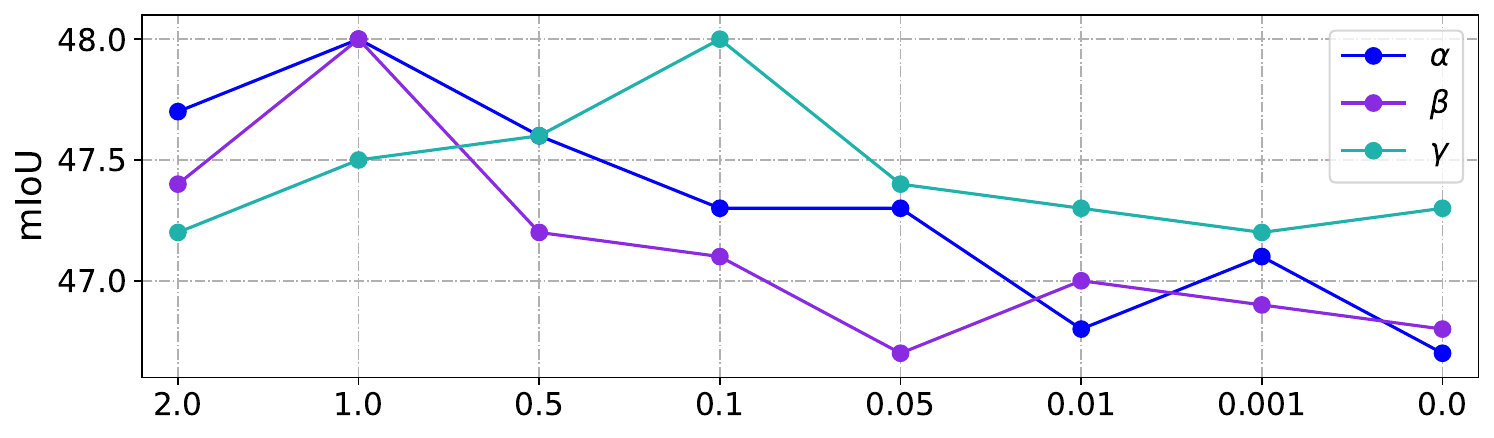}
\caption{Ablation study for volume regularization, anchor guidance and convex guarantee of SimT.}
\label{fig:ablate}
\end{figure}

\textbf{Ablation Experiments.} To investigate effects of individual components within SimT, we design ablation experiments under UDA setting (GTA5$\rightarrow$Cityscapes), as shown in Table \ref{ablation}. In particular, we validate three proposed regularizers, \ie volume regularization, anchor guidance and convex guarantee, and the auxiliary loss devised for anchor guidance is also ablated to validate its effectiveness. We adopt AdaptSegNet \cite{tsai2018learning} as the black-box model to generate pseudo labels for target domain data. It is observed that incorporating SimT ($row$ 11) to model mixed closed-set and open-set noise distributions can correct supervision signals on target data and obtain the performance gain with 5.6\% increment in mIoU. Ablating each proposed component ($rows$ 7-10) induces the performance degradation. It is because that the designed regularizers are complementary and combining all regularizers with properly chosen coefficients (\ie $\alpha=1.0$, $\beta=1.0$, $\gamma=0.1$) can reach a trade-off to better approximate the true SimT, as shown in Figure \ref{fig:ablate}. 

\textbf{Effectiveness of SimT.} We then compare SimT with four typical self-training based UDA models \cite{zou2019confidence, zou2018unsupervised, zheng2021rectifying, guo2021metacorrection}. For a fair comparison, all competed methods use AdaptSegNet to derive pseudo labels. As listed in Table \ref{ablation}, the proposed SimT ($row$ 11) is superior to other self-training methods, including entropy minimization \cite{zou2019confidence} ($row$ 3), threshold \cite{zou2018unsupervised} ($row$ 4) and uncertainty \cite{zheng2021rectifying} ($row$ 5) based sample selection, MetaCorrection based rectification \cite{guo2021metacorrection} ($row$ 6), yielding increments of 2.9\%, 3.6\%, 1.9\%, 0.7\% mIoU. With entropy minimization, considerable noisy labels inevitably lead to unsatisfactory adaptation. 
Although self-training methods in \cite{zheng2021rectifying, zou2018unsupervised} discard noisy samples and may filter out open-set label noises, but lose useful information in those discarded pixels, especially when they are around the decision boundary. Both MetaCorrection \cite{guo2021metacorrection} and SimT distill effective information from all samples. Comparing with MetaCorrection that only models closed-set noise distribution, the proposed SimT additionally ameliorates open-set noise issues and shows the superior performance.

\begin{figure}
\centering
\vspace{+0.2cm}
\includegraphics[width=82mm]{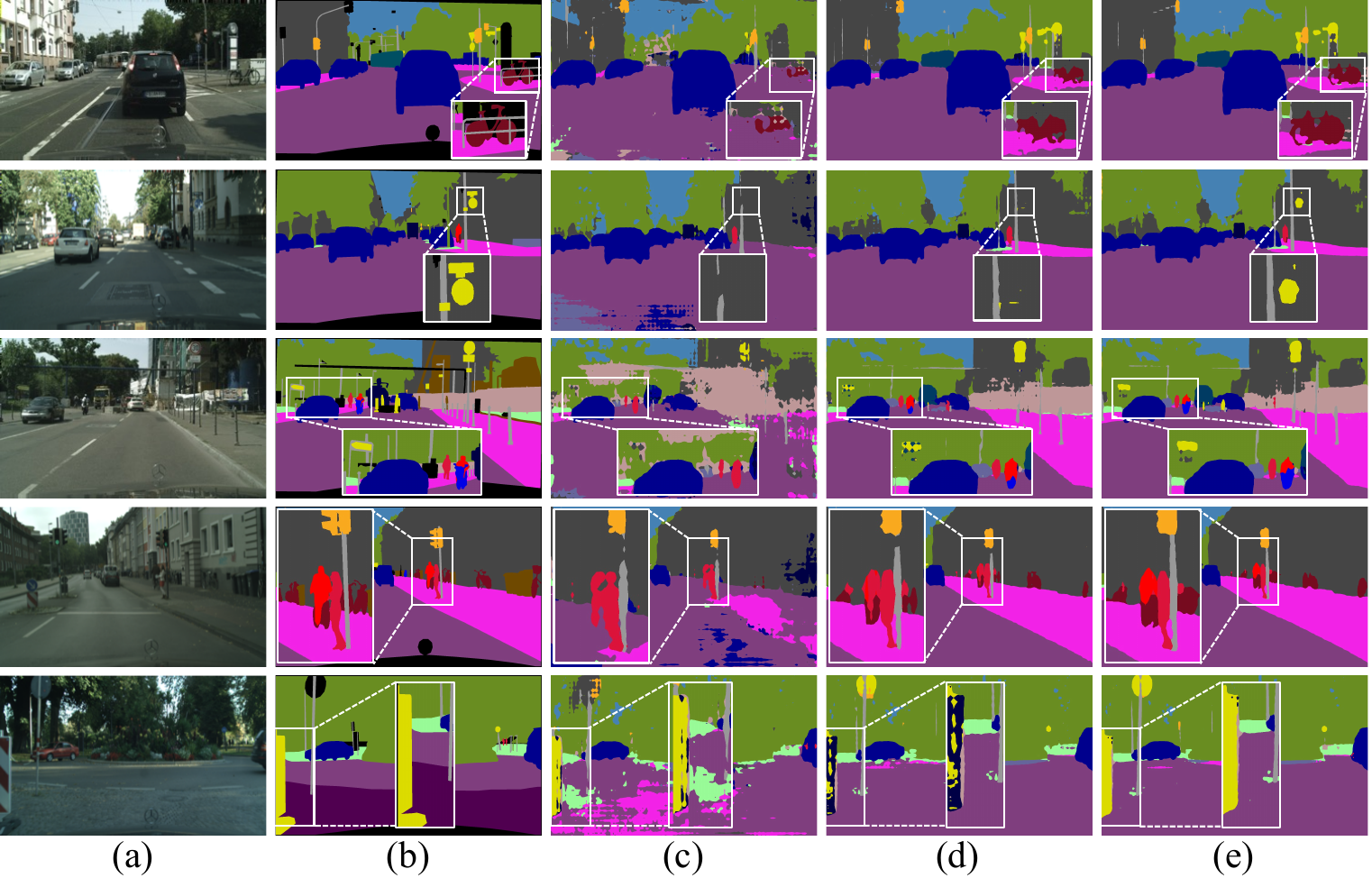}
\caption{Qualitative results of UDA in GTA5$\rightarrow$Cityscapes scenario. (a) Target image,  (b) Ground truth, Predictions from (c) source only model, (d) BAPA-Net \cite{liu2021bapa}, (e) ours (SimT).}
\label{fig:seg}
\end{figure}

\begin{figure}
\centering
\includegraphics[width=82mm]{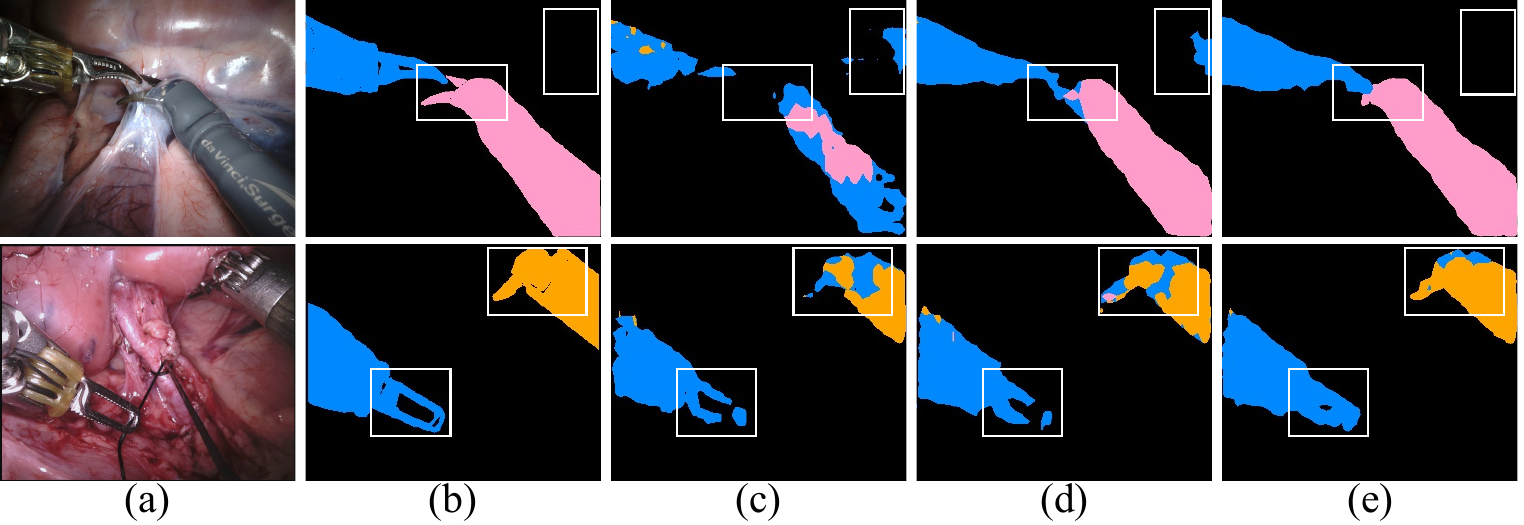}
\caption{Qualitative results of UDA in Endovis17$\rightarrow$Endovis18 scenario. (a) Target image,  (b) Ground truth, Predictions from (c) source only model, (d) IGNet \cite{liu2021ignet}, (e) ours (SimT).}
\label{fig:seg_instru}
\end{figure}

\textbf{Robustness to Various Types of Noise.}
To explore the robustness of our SimT, we further evaluate it under different types of label noises. To be specific, we adopt four UDA models, including AdaptSegNet \cite{tsai2018learning}, LTIR \cite{kim2020learning}, DSP \cite{gao2021dsp} and BAPA-Net \cite{liu2021bapa}, as black-box models to generate noisy pseudo labels. Then SimT is applied to mitigate the closed-set and open-set noise problem within pseudo labels. As illustrated in Table \ref{ablation}, these four UDA models have significantly improved performance with the proposed SimT.

\begin{figure}
\centering
\includegraphics[width=82mm]{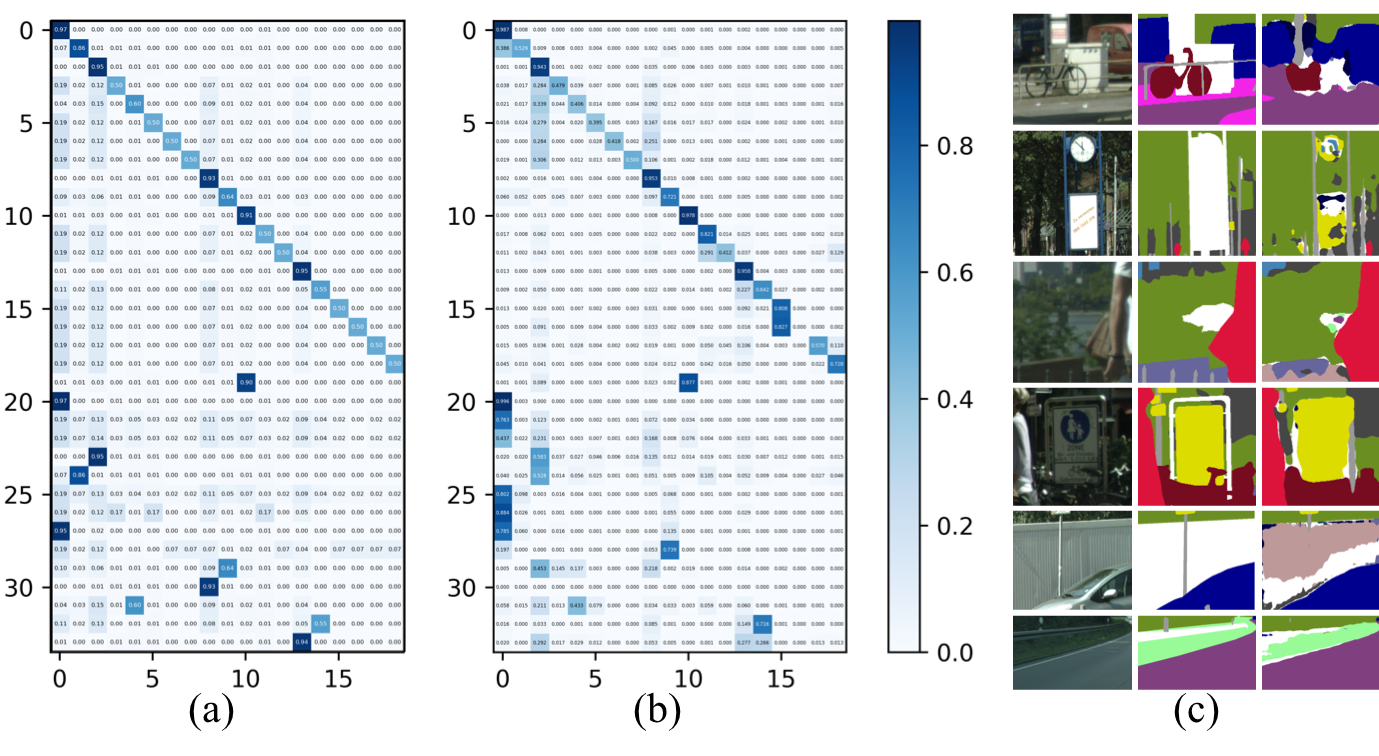}
\caption{Visualization results. (a) The learned SimT, (b) confusion matrix of pseudo labels, (c) open-set class visualization.}
\label{fig:tsne}
\end{figure}

\subsection{Visualization Results}
\textbf{Segmentation Visualization.} 
As illustrated in Figure \ref{fig:seg}, we provide some typical qualitative segmentation results of target data in GTA5$\rightarrow$Cityscapes scenario. Compare with the baseline BAPA-Net \cite{liu2021bapa}, the proposed SimT shows superior performance in identifying small-scale objectives (\eg, `traffic sgn', $rows$ 2, 3, 5) and differentiating confused class samples (\eg, `rider', `person' and `bike', $rows$ 1, 4). In practice, we observe that label noises in pseudo-labeled target data tend to appear in the minor categories and ambiguous categories, leading to the unsatisfactory performance of baselines. On the contrary, the proposed SimT corrects supervision signals for target data and alleviates the noise issues. We also illustrate qualitative results of target Endovis18 data under UDA setting, as shown in Figure \ref{fig:seg_instru}. 

\textbf{SimT Visualization.} Inspired by the fact that confusion matrix is a crude estimator of the true NTM \cite{zhang2021learning}, we visualize the learned SimT and the confusion matrix of pseudo labels for reference, as illustrated in Figure \ref{fig:tsne} (a, b). It is obvious that the optimized SimT is similar to confusion matrix, demonstrating the proposed SimT can model the closed-set and open-set noise distributions and generate corrected supervision signals for pseudo-labeled target data.

\textbf{Open-set Class Visualization.} 
We further visualize detected open-set class regions, which are denoted by `white' areas in Figure \ref{fig:tsne} (c). From left to right, they are input image, ground truth and clean class posterior prediction, respectively. We observe the proposed method is able to detect open-set class pixels and inherently has the capability of alleviating open-set label noise issue in target data. 

\section{Conclusion}
In this paper, we study a general and practical DA semantic segmentation setting where only pseudo-labeled target data is accessible through a black-box model. SimT is advanced to model mixed close-set and open-set noise distributions within pseudo-labeled target data. We formulate columns of SimT to make up a simplex and interpret it through the computational geometry analysis. Based on the geometry analysis, we devise three regularizers, \ie volume regularization, anchor guidance, convex guarantee, to approximate the true SimT, which is further utilized to correct noise issues in pseudo labels and enhance the generalization ability of segmentation model on the target domain data. Extensive experimental results demonstrate that the proposed SimT can be flexibly plugged into existing UDA and SFDA methods to boost the performance. In the future, we plan to introduce open-set class prior to enable the semantic separation inner detected open-set regions. 

\appendix
\section{Supplementary}

In the supplementary material, we first summarize notations used in the manuscript. Then we provide extensive implementation details with additional qualitative and quantitative performance analysis. Towards reproducible research, we will release our code and optimized network weights. This supplementary is organized as follows:

\begin{itemize}
 \item Section \ref{Notations}: Notations 
  
 \item Section \ref{Implement}: Implementation details 
 
 	-- Experimental settings (Sec. \ref{setting})
	
	-- Reparameterization of SimT (Sec. \ref{SimT})
	
	-- Reparameterization of weighting matrix (Sec. \ref{weightM}) 

 \item Section \ref{exp}: Experimental results
 
  	-- Analysis on open-set classes (Sec. \ref{OSC})

  	-- Segmentation visualization of SFDA (Sec. \ref{sfda})

 \end{itemize}
 
\subsection{Notations}\label{Notations}
We summarize the notations utilized in the manuscript, as listed in Table \ref{table:Symbols}. 

\begin{table}[t]
\centering
\caption{Notation Table.}
\label{table:Symbols}
\scalebox{0.83}{\begin{tabular}{l   l}
\toprule[1pt]
Symbol&Description\\
\hline
\xrowht{7pt}
$\boldsymbol{T}$&SimT\\

\xrowht{7pt}
$\boldsymbol{x}$&Pixel\\

\xrowht{7pt}
$\widetilde{\boldsymbol{y}}$&Pseudo label of pixel $\boldsymbol{x}$ \\

\xrowht{7pt}
$\boldsymbol{y}$&Ground truth label of pixel $\boldsymbol{x}$ \\

\xrowht{7pt}
$X_{t}$&Target domain image\\

\xrowht{7pt}
$\widetilde{Y}_{t}$&Pseudo label of image $X_{t}$ \\

\xrowht{7pt}
$Y_{t}$&Ground truth label of image $X_{t}$ \\

\xrowht{7pt}
$\mathcal{X}_{\mathcal{T}}$&Set of target domain images \\

\xrowht{7pt}
$\mathcal{Y}_{\mathcal{T}}$&Set of target domain pseudo segmentation labels \\

\xrowht{7pt}
$p(\widetilde{\boldsymbol{y}} \mid \boldsymbol{x})$&Noisy class posterior probability \\

\xrowht{7pt}
$p({\boldsymbol{y}} \mid \boldsymbol{x})$&Clean class posterior probability \\

\xrowht{7pt}
$f_{b}(\cdot)$&Balck-box model \\

\xrowht{7pt}
$f(\cdot)_\mathbf{w}$&Segmentation model parameterized by $\mathbf{w}$ \\

\xrowht{7pt}
$f(\cdot)_{\mathbf{w}^{fixed}}$&Warm-up model parameterized by $\mathbf{w}^{fixed}$ \\

\xrowht{7pt}
$\boldsymbol{x}^{c}$&Anchor point of class $c$ \\

\xrowht{7pt}
$\boldsymbol{x}_{k}$&Confident closed-set pixel \\

\xrowht{7pt}
$\widetilde{\boldsymbol{y}}_{k}$&Label of confident closed-set pixel $\boldsymbol{x}_{k}$ \\

\xrowht{7pt}
$\boldsymbol{x}_{u}$&Confident open-set pixels \\

\xrowht{7pt}
$\widetilde{\boldsymbol{y}}_{u}$&Label of confident open-set pixel $\boldsymbol{x}_{u}$ \\

\xrowht{7pt}
$X_{\mathcal{K}}$&Set of confident closed-set pixels \\

\xrowht{7pt}
$X_{\mathcal{U}}$&Set of confident open-set pixels \\

\xrowht{7pt}
$\boldsymbol{u}$&Weighting matrix\\

\xrowht{7pt}
$\mathcal{L}_{ST}$&Self-training loss\\

\xrowht{7pt}
$\mathcal{L}_{LC}$&Loss correction\\

\xrowht{7pt}
$\mathcal{L}_{Volume}^{SimT}$&Volume regularization \\

\xrowht{7pt}
$\mathcal{L}_{Anchor}^{SimT}$&Anchor guidance \\

\xrowht{7pt}
$\mathcal{L}_{Convex}^{SimT}$&Convex guarantee \\

\xrowht{7pt}
$\alpha$, $\beta$, $\gamma$&Regularization coefficients\\

\bottomrule[1pt]
\end{tabular}}
\end{table}

\subsection{Implementation details}\label{Implement}

\subsubsection{Experimental settings}\label{setting}

\textbf{Hyper-parameters in Endovis17$\rightarrow$Endovis18 scenario.} 
We adopt polynomial learning rate scheduling to optimize the feature extractor with the initial learning rate of 1e-4, while it is set as 1e-3 for the optimization of classifier and SimT. The batch size is set as 8, and the maximum epoch number is 30. Hyper-parameters of $n$, $\lambda$, $\alpha$, $\beta$, $\gamma$ are set as 5, 0.1, 1.0, 1.0, 0.1 in our implementation. 

\textbf{Detailed training and inference procedure.} 
We adopt DeepLab-v2 \cite{chen2017deeplab} backbone with encoder of ResNet-101 \cite{he2016deep} as our segmentation model. Given the pseudo-labeled target domain data derived from a given black-box model $f_{b}(\cdot)$, we calculate the class distribution of pseudo labels $\boldsymbol{C}$. During training phase, we first warm up the whole segmentation model to obtain $f(\cdot)_{\mathbf{w}^{fixed}}$ with $C$-way output probabilities using the generated pseudo-labeled target domain data. The warm-up model is utilized to produce noisy class posteriors for anchor points and derive confidence score for each pixel. The classifier of segmentation model $f(\cdot)_{\mathbf{w}^{fixed}}$ is then extended to output $(C+n)$-way clean class posterior probabilities, and the extended model is denoted as $f(\cdot)_{\mathbf{w}}$. Incorporating SimT, we only update $conv3$ and  $conv4$ layer in feature extractor and $(C+n)$-way classifier through gradient decent derived from corrected loss. During inference phase, we obtain final predictions from first $C$-way of the extended model $f(\cdot)_{\mathbf{w}}$ directly. 

\subsubsection{Reparameterization of SimT ($\boldsymbol{T}$)}\label{SimT}
To make SimT ($\boldsymbol{T}$) differentiable and satisfy the condition of $\boldsymbol{T} \in [0, 1]^{(C+n)\times C}$, $\sum_{k=1}^{C} \boldsymbol{T}_{jk}=1$, we utilize the reparameterization method, and the code is shown in Listing \ref{lstlisting:SimT}. Specifically, we randomly initialize a matrix $\boldsymbol{U} \in \mathbb{R}^{(C+n)\times C}$, as in lines 8-14. To preserve the diagonally dominant property of closed-set part of SimT ($\boldsymbol{T}_{1:C,:}$), the diagonal prior $\boldsymbol{I}$ is introduced in line 16, which has been widely used in the literature of NTM estimation \cite{patrini2017making, li2021provably}. Considering segmentation model tends to classify samples into majority categories, the class distribution of pseudo labels $\boldsymbol{C}$ is involved in lines 18-20. Incorporating these prior informations, we obtain $\boldsymbol{V} = \boldsymbol{C} \cdot \sigma (\boldsymbol{U}) + \boldsymbol{I}$ in lines 23-24, where $\sigma(\cdot)$ is the sigmoid function to avoid negative value in SimT. Then we do the normalization of $\boldsymbol{T}_{jk} = \frac{\boldsymbol{V}_{jk}}{\sum_{k=1}^{C} \boldsymbol{V}_{jk}}$ to derive SimT ($\boldsymbol{T}$) in line 25. Since both sigmoid function and normalization operation are differentiable, SimT can be updated through gradient descent on $\boldsymbol{U}$.

\vspace{+0.2cm}
\begin{lstlisting}[language=Python, caption=Reparameterization of SimT ($\boldsymbol{T}$), label={lstlisting:SimT}]
import torch
import torch.nn as nn
import torch.nn.functional as F
import numpy as np
class SimT(nn.Module):
    def __init__(self, num_classes, open_classes=0):
        super(SimT, self).__init__()
        T = torch.ones(num_classes+open_classes, num_classes)
        
        self.register_parameter(name='NTM', param=nn.parameter.Parameter(torch.FloatTensor(T)))
        
        self.NTM  # U
        
        nn.init.kaiming_normal_(self.NTM, mode='fan_out', nonlinearity='relu')
        
        self.Identity_prior = torch.cat([torch.eye(num_classes, num_classes), torch.zeros(open_classes, num_classes)], 0) # I
        
        Class_dist = np.load('../ClassDist/AdaptSegNet_CD.npy') # C
        
        self.Class_dist = torch.FloatTensor(np.tile(Class_dist, (num_classes + open_classes, 1)))
        
    def forward(self):
        T = torch.sigmoid(self.NTM).cuda()
        T = T.mul(self.Class_dist.cuda().detach()) + self.Identity_prior.cuda().detach() # V
        T = F.normalize(T, p=1, dim=1) # SimT
        return T
\end{lstlisting}

\subsubsection{Reparameterization of weighting matrix ($\boldsymbol{u}$)}\label{weightM}
In the proposed convex guarantee of SimT, we introduce a weighting matrix $\boldsymbol{u} \in [0, 1]^{(C+n)\times (C+n)}$ with constraints of $\boldsymbol{u}_{j,k=j} = -1$ and $\sum_{k}\boldsymbol{u}_{j,k\neq j} =1$. To make the weighting matrix $\boldsymbol{u}$ differentiable and satisfy its constraints, we utilize the reparameterization method, as shown in Listing \ref{lstlisting:u}. To be specific, we uniformly initialize a matrix $\boldsymbol{W} \in \mathbb{R}^{(C+n)\times (C+n)}$ except diagonal entries in lines 9-15. The softmax operator is introduced to ensure the summation of non-diagonal entries in each row of $\boldsymbol{u}$ to be 1, as shown in line 24. Diagonal entries are all detached and set as -1, as in lines 17-25. Since the softmax operation is differentiable, the weighting matrix $\boldsymbol{u}$ can be updated through gradient descent on $\boldsymbol{W}$.  

\vspace{+0.2cm}
\begin{lstlisting}[language=Python, caption=Reparameterization of weighting matrix ($\boldsymbol{u}$), label={lstlisting:u}]
import torch
import torch.nn as nn
import torch.nn.functional as F
import numpy as np
class sig_u(nn.Module):
    def __init__(self, num_classes, open_classes=0):
        super(sig_u, self).__init__()
        
        self.classes = num_classes+open_classes
        
        init = 1./(self.classes-1.)

        self.register_parameter(name='weight', param=nn.parameter.Parameter(init*torch.ones(self.classes, self.classes)))

        self.weight # W

        self.identity = torch.zeros(self.classes, self.classes) - torch.eye(self.classes)
        
    def forward(self):
        ind = np.diag_indices(self.classes)
        with torch.no_grad():
            self.weight[ind[0], ind[1]] = -10000. * torch.ones(self.classes).detach()

        w = torch.softmax(self.weight, dim = 1).cuda()
        weight = self.identity.detach().cuda() + w # u
        return weight
\end{lstlisting}

\subsection{Experimental results}\label{exp}

\subsubsection{Analysis on open-set classes}\label{OSC}

\begin{figure}
\centering
\includegraphics[width=84mm]{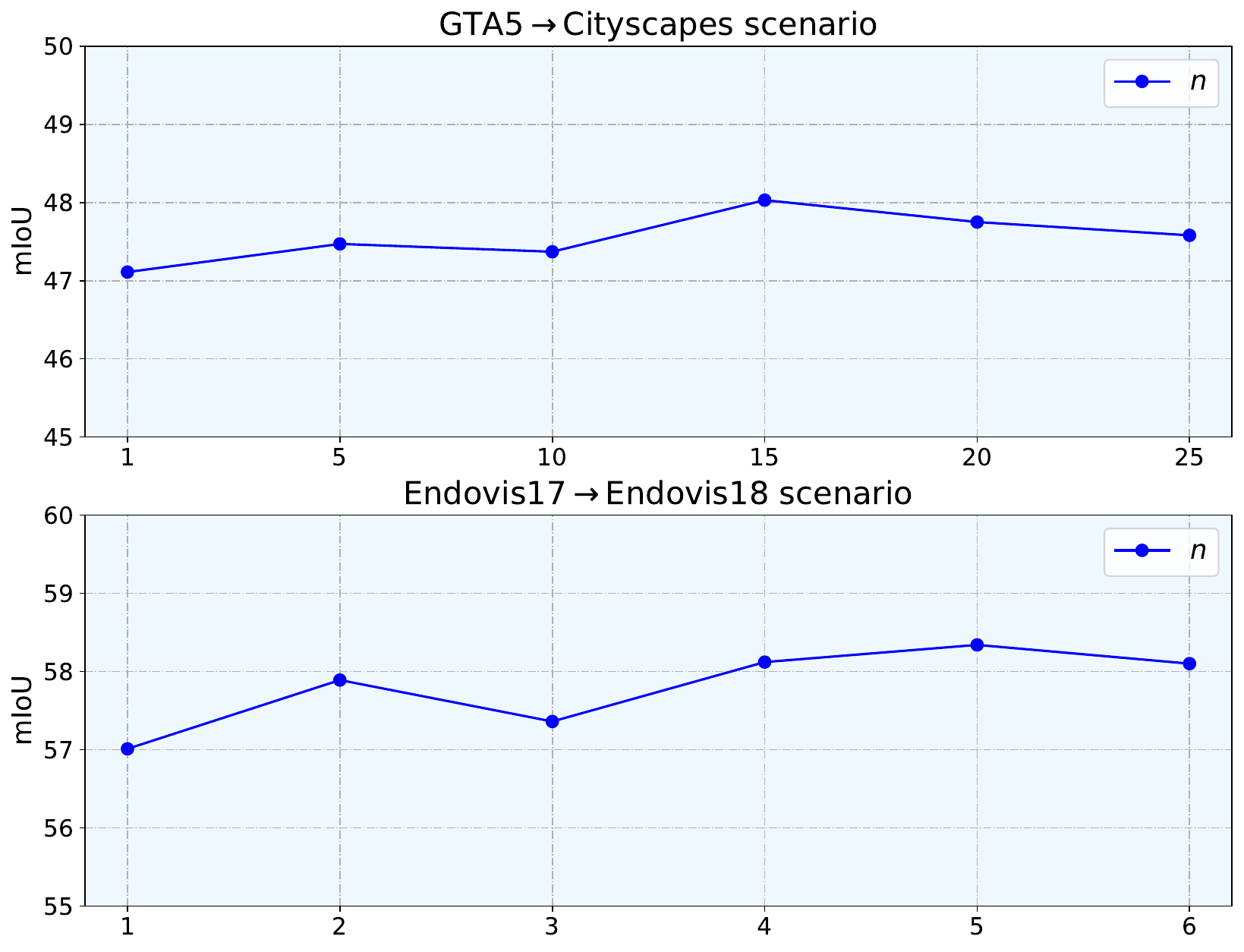}
\caption{Qualitative results of SFDA in Endovis17$\rightarrow$Endovis18 scenario. (a) Target image,  (b) Ground truth, Predictions from (c) source only model, (d) IGNet \cite{liu2021ignet}, (e) ours (SimT).}
\label{fig:SimT_n}
\end{figure}

In UDA of GTA5$\rightarrow$Cityscapes scenario, the compatible label set shared between GTA5 \cite{richter2016playing} and Cityscapes \cite{cordts2016cityscapes} includes 19 classes, \ie `road', `sidewalk', `building', `wall', `fence', `pole', `traffic-light', `traffic-sign', `vegetable', `terrain', `sky', `person', `rider', `car', `truck', `bus', `train', `motor', `bike'. In target domain (Cityscapes), 15 open-set classes, including `ego vehicle', `rectification border', `out of roi', `static', `dynamic', `ground', `parking', `rail track', `guard rail', `bridge', `tunnel', `polegroup', `caravan', `trailer', `license plate' are unknown in source domain (GTA5)\footnote{\url{https://github.com/mcordts/cityscapesScripts/blob/master/cityscapesscripts/helpers/labels.py}}. In the proposed SimT, $n$ is a hyper-parameter that indicates the potential open-set class number, and we set $n=15$ to implicitly model the diverse semantics within open-set classes. Considering the prior of open-set class number is not available in real-word deployment, we tune it in the range of \{1, 5, 10, 15, 20, 25\} and show the influence of open-set class number $n$, as illustrated in the upper row of Figure \ref{fig:SimT_n}. In this experiment, we adopt AdaptSegNet \cite{tsai2018learning} as the black-box model to generate pseudo labels for target domain data. It is observed that $n=1$ shows the inferior performance. This result validates that multiple open-set ways of classifier capacitate the segmentation model to encode the diverse feature representations within open-set regions. Moreover, we also observe that the performance of segmentation model remains stable across a wide range of $n$, indicating the robustness of the proposed SimT.

In UDA of Endovis17$\rightarrow$Endovis18 scenario, the compatible label set of instrument types shared between Endovis17 \cite{allan20192017} and Endovis18 \cite{allan20202018} includes 3 classes, \ie `scissor', `needle driver', `forceps'. In target domain (Endovis18), 3 open-set instrument type classes, including `ultrasound probe', `suction instrument', `clip applier' are as the unknown classes in source domain (Endovis17). We tune the open-set class number $n$ in the range of \{1, 2, 3, 4, 5, 6\}, and the comparison results are  shown in the lower row of Figure \ref{fig:SimT_n}. In this experiment, the black-box model is borrowed from IGNet \cite{liu2021ignet} to generate pseudo labels for target domain data. It is clear that the verification performance of segmentation model with various $n$ settings remains stable within a wide range.

\subsubsection{Segmentation visualization of SFDA}\label{sfda}

\begin{figure}
\centering
\includegraphics[width=84mm]{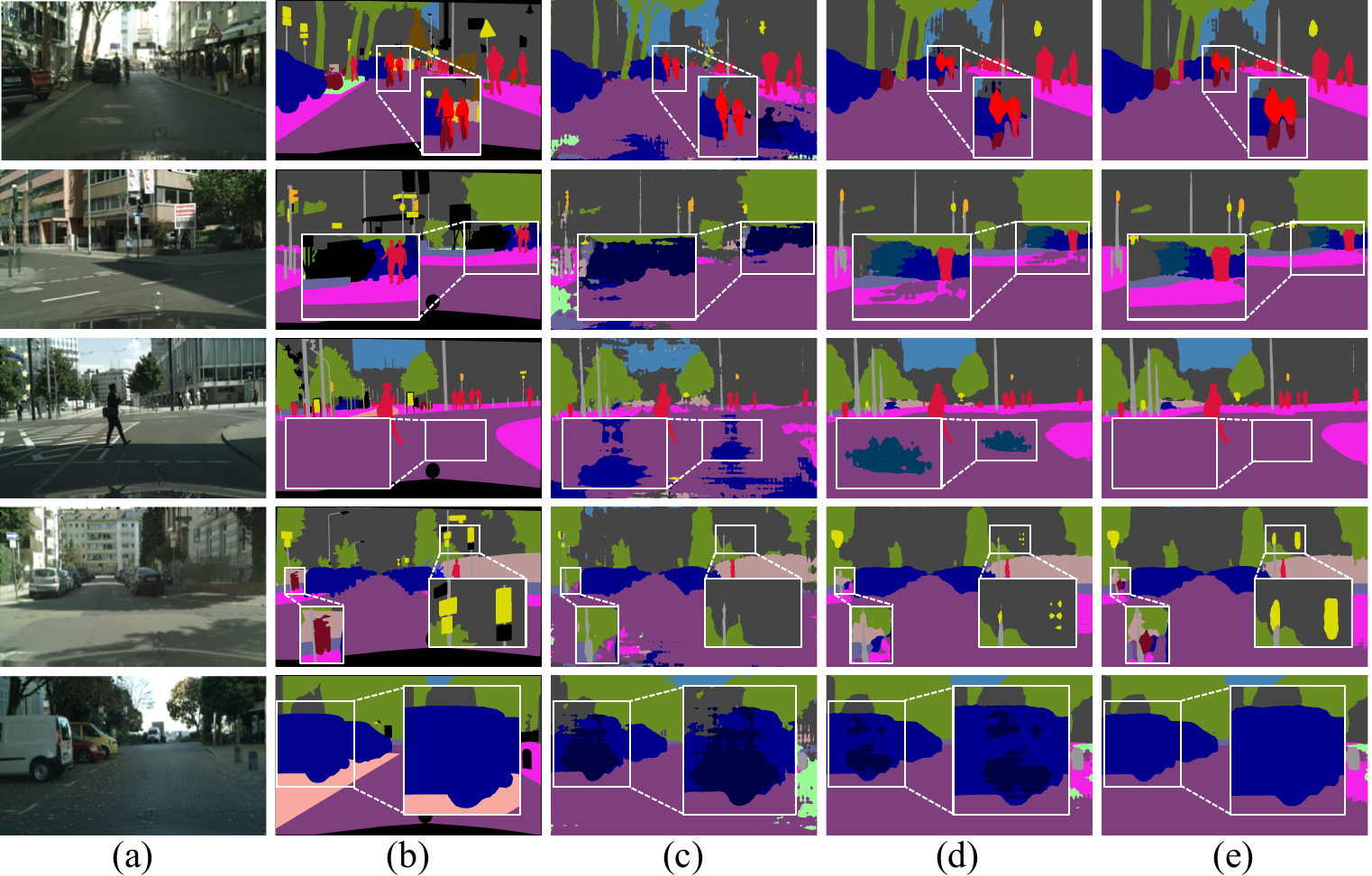}
\caption{Qualitative results of SFDA in GTA5$\rightarrow$Cityscapes scenario. (a) Target image,  (b) Ground truth, Predictions from (c) source only model, (d) SFDASeg \cite{kundu2021generalize}, (e) ours (SimT).}
\label{fig:seg_cv}
\end{figure}

\begin{figure}
\centering
\includegraphics[width=84mm]{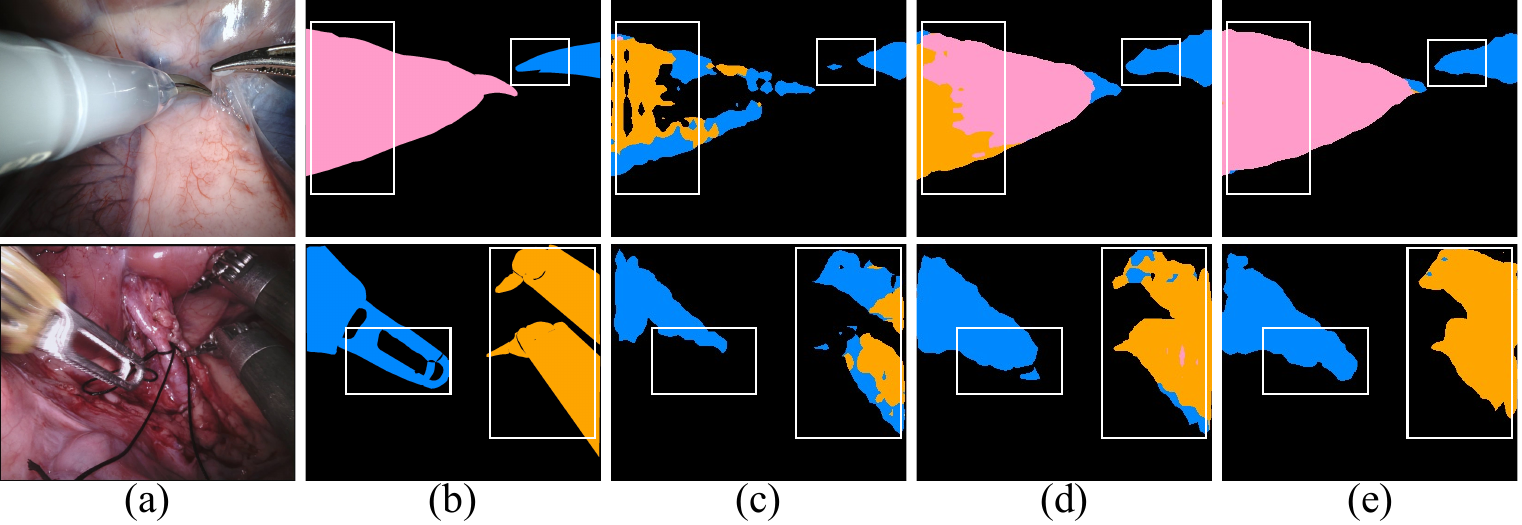}
\caption{Qualitative results of SFDA in Endovis17$\rightarrow$Endovis18 scenario. (a) Target image,  (b) Ground truth, Predictions from (c) source only model, (d) SFDASeg \cite{kundu2021generalize}, (e) ours (SimT).}
\label{fig:seg_instru}
\end{figure}

In Figure \ref{fig:seg_cv}, we present segmentation results on the SFDA semantic segmentation task in GTA5$\rightarrow$CityScapes scenario. As shown in the figure, the source only model (c) produces the worst segmentation results, since the extracted features for target data are not discriminative enough. The baseline SFDA model (d) \cite{kundu2021generalize} obtains more precise segmentation predictions in comparison to the source only model, but is error-prone in some ambiguous categories (\eg, `rider' and `bike', `road' and `sidewalk') and small-scale objects (\eg, `traffic sign'). In the results of our SimT approach, these mistakes are effectively mitigated, resulting in more reasonable segmentation predictions. We conjecture the reason is that our method can adaptively model the noise distribution of pseudo labels in target domain and learn the discriminative feature representation of target data with the corrected supervision signals.

We provide segmentation results on the SFDA semantic segmentation task of Endovis17$\rightarrow$Endovis18 in Figure \ref{fig:seg_instru}. The qualitative results show that without adaptation (source only model), it is difficult to correctly identify the surgical instruments due to the limited discriminative capability on the target data. The predicted instrument regions are fragmentary, deteriorating the segmentation performance. Compared with the baseline SFDA model (d) \cite{kundu2021generalize}, the proposed approach (e) generates more semantically meaningful segmentation results, demonstrating the superior property of SimT in alleviating the noise issues in SFDA task.

{\small
\bibliographystyle{ieee_fullname}
\bibliography{egbib}
}

\end{document}